\documentclass[10pt,twocolumn]{article}

\usepackage[T1]{fontenc}
\usepackage[utf8]{inputenc}
\input{glyphtounicode}
\usepackage{mathptmx}
\usepackage[scaled=0.92]{helvet}
\usepackage{courier}
\usepackage[margin=0.63in, top=0.62in, bottom=0.64in, columnsep=0.22in]{geometry}
\usepackage[protrusion=true,expansion=false]{microtype}
\usepackage{amsmath,amssymb,amsfonts}
\usepackage{array}
\usepackage{booktabs}
\usepackage{tabularx}
\usepackage{graphicx}
\usepackage{xcolor}
\definecolor{runelabteal}{HTML}{0F766E}
\usepackage{enumitem}
\usepackage{titlesec}
\usepackage{url}
\usepackage[colorlinks=true,linkcolor=runelabteal,citecolor=runelabteal,urlcolor=runelabteal,pdfencoding=auto,pdftitle={Train Once, Transfer Where? Better Alignment Does Not Mean Better Probe Transfer},pdfauthor={Engin Deniz Alpman and H\"useyin Oktay Altun},pdfkeywords={cross-model probe transfer, linear probes, representational alignment, Orthogonal Procrustes, mean-shift probes, LLM safety monitoring, prompt injection detection}]{hyperref}

\renewcommand{\arraystretch}{1.05}
\setlist{nosep,leftmargin=1.25em}
\titleformat{\section}{\normalfont\large\bfseries\raggedright}{\thesection}{0.55em}{}
\titleformat{\subsection}{\normalfont\normalsize\bfseries\raggedright}{\thesubsection}{0.55em}{}
\titlespacing*{\section}{0pt}{6.5pt plus 1.2pt minus 1.0pt}{2.5pt plus 0.5pt}
\titlespacing*{\subsection}{0pt}{5.0pt plus 1.0pt minus 0.8pt}{1.8pt plus 0.4pt}
\titlespacing*{\paragraph}{0pt}{3.2pt plus 0.6pt}{0.45em}

\makeatletter
\long\def\@makecaption#1#2{%
  \vskip\abovecaptionskip
  \sbox\@tempboxa{\small\textbf{#1:} #2}%
  \ifdim \wd\@tempboxa >\hsize
    {\small\textbf{#1:} #2\par}
  \else
    \global \@minipagefalse
    \hb@xt@\hsize{\hfil\box\@tempboxa\hfil}%
  \fi
  \vskip\belowcaptionskip}
\makeatother

\newcommand{\stackcell}[2][l]{\begin{tabular}[c]{@{}#1@{}}#2\end{tabular}}

\newcommand{\codeRepoNote}{Code, verified JSON metrics, and plotting scripts are available at \url{https://github.com/RuneLabAI/linear-probe-transfer}.}
\newcommand{\preregCommitNote}{; \texttt{v9}/\texttt{v9b} consolidation rules were committed prior to execution at git commits \texttt{4169438} and \texttt{c2260c0}}

\begin{document}

\twocolumn[{%
\vspace*{-14pt}
\noindent\href{https://runelab.ai/en}{\includegraphics[height=11pt]{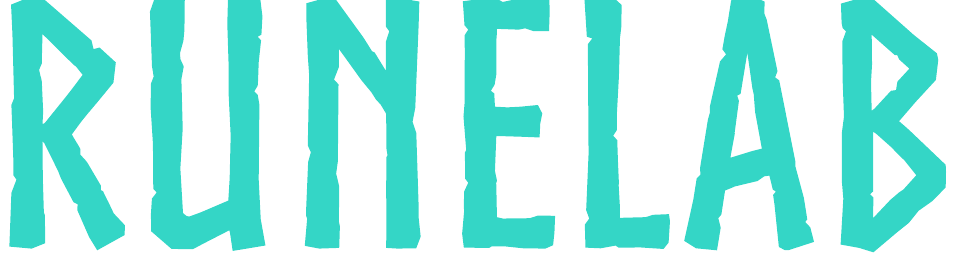}}\hfill{\footnotesize\textcolor{gray}{Preprint $\cdot$ December 2026}}\\[3pt]
\begin{center}
  \vspace{-11pt}
  \noindent\rule{\textwidth}{2.2pt}\\[4pt]
  {\Large\bfseries Train Once, Transfer Where?\\Better Alignment Does Not Mean Better Probe Transfer\par}
  \vspace{3.5pt}
  \noindent\rule{\textwidth}{1pt}\\[5pt]
  \begin{minipage}[t]{0.48\textwidth}
    \centering
    \textbf{Engin Deniz Alpman}$^{\dagger}$\\[1pt]
    {\small \href{https://runelab.ai/en}{RuneLab.ai} \& Institute for Data Science and Artificial Intelligence,\\
    Bo\pdfliteral{/Span<</ActualText<FEFF011F>>>BDC}\u{g}\pdfliteral{EMC}azi\c{c}i University, Istanbul, T\"urkiye\\
    \href{mailto:engin@runelab.ai}{\texttt{engin@runelab.ai}}}
  \end{minipage}
  \hfill
  \begin{minipage}[t]{0.48\textwidth}
    \centering
    \textbf{H\"useyin Oktay Altun}\\[1pt]
    {\small Institute for Data Science and Artificial Intelligence,\\
    Bo\pdfliteral{/Span<</ActualText<FEFF011F>>>BDC}\u{g}\pdfliteral{EMC}azi\c{c}i University, Istanbul, T\"urkiye\\
    \texttt{\href{mailto:oktay.altun@bogazici.edu.tr}{oktay.altun@bogazici.edu.tr}}}
  \end{minipage}
  \vspace{3pt}
\end{center}
\begin{quote}
\small
\begin{center}\textbf{Abstract}\end{center}
\vspace{-5pt}
Linear probes on frozen language model activations cheaply monitor safety, truthfulness, and prompt injection, yet require retraining across model upgrades. Can read-only probes transfer across models from unlabeled text? Across six models from four families (Qwen, Gemma, Llama~3.1, OLMo~3; 30 pairs), probes mapped via 600-sentence Orthogonal Procrustes rotations retain 94\%--99.5\% matched native above-chance AUROC on core input-property tasks, beat full multi-layer global Relative Representations by $+0.11$ matched logistic and $+0.19$ matched mean-shift AUROC, and fold into one dot product with negligible memory (16\,KB/head). Transfer degrades on behavioral sycophancy (reflecting label disagreement and direction loss), long-context injections, and adaptive evasion. Across four core tasks (\texttt{hh-rlhf}, \texttt{XSTest}, \texttt{TruthfulQA}, \texttt{MATH-500}), two findings stand out. First, zero-hyperparameter mean-shift probes transfer best, matching or beating source-tuned logistic regression. Second, better anchor fit does not aid transfer: whitened, Ridge, Kernel Ridge, and neural maps fit anchors better up to 20,000 sentences yet trail Procrustes because anisotropic shrinkage distorts target directions $3\times$--$7\times$ and lowers mean test SNR. Balanced four-domain anchors lift \texttt{deepset} OOD transfer ($+0.10$), unlabeled in-domain anchors further raise four-task transfer ($+0.02$), and multi-source averaging closes 38\%--46\% of non-ceiling gaps without target labels.

\end{quote}
\vspace{1pt}
}]

\begingroup
\renewcommand{\thefootnote}{}%
\footnotetext{\raggedright $^{\dagger}$Corresponding author (\href{mailto:engin@runelab.ai}{\texttt{engin@runelab.ai}}). \href{https://runelab.ai/research/linear-probe-transfer}{Project page} (\url{https://runelab.ai/research/linear-probe-transfer}) and \href{https://github.com/RuneLabAI/linear-probe-transfer}{GitHub} (\url{https://github.com/RuneLabAI/linear-probe-transfer}).}%
\endgroup

\section{Introduction}
\label{sec:intro}

Deploying language models in production requires screening prompts for safety violations, prompt injections, and task difficulty before generation begins. Linear probes on frozen prefill activations avoid auxiliary encoder models, yet backbone upgrades across vendors, scales, or architectures typically demand fresh target labels. When original labeled texts remain accessible, native retraining on the new model is inexpensive and usually preferable. Cross-model probe transfer serves zero-label cold-start upgrades when source corpora were deleted, cannot cross privacy or compliance boundaries, or belong to an external vendor during rapid rollouts.

We ask whether linear probes trained on one model can transfer onto a frozen target model using unlabeled neutral text. We restrict attention to read-only monitoring rather than writing rotated states or weights back into the forward pass, which failed pre-registered criteria in preliminary write-path experiments.

Representational alignment studies shared coordinates across networks, from model stitching~\cite{lenc2015understanding,bansal2021stitching} and the Platonic Representation Hypothesis~\cite{huh2024platonic} to Orthogonal Procrustes~\cite{schonemann1966procrustes}, which outperforms unconstrained linear maps in cross-lingual word-embedding alignment~\cite{xing2015normalized,conneau2018word}. Cheng et al.~\cite{cheng2026platoskey} (\textit{Plato's Key}) showed that Orthogonal Procrustes transfers factual representations across models. Within single models, Marks and Tegmark~\cite{marks2024geometry} observed that mass-mean probes generalize out of distribution better than logistic regression, and Lee et al.~\cite{lee2026hallucination} showed that hallucination signals behave largely as a mean shift.

For cross-model monitoring, Moschella et al.~\cite{moschella2023relative} introduced Relative Representations, and Bakker and Louizos~\cite{bakker2026relative} aggregated angular distances across all transformer layers; we benchmark both their full multi-layer construction and fixed-depth cosine Relative Representations against folded Procrustes. Concurrent work compares covariance-adjusted probing against mean-difference steering across models~\cite{sadeghi2026probing}, projects behavioral axes into shared anchor spaces~\cite{kim2026universality}, transfers safety directions to visual generators~\cite{poppi2026safety}, analyzes noise sensitivity in whitened inner products~\cite{svendsen2026improving}, aligns concept atlases~\cite{puri2026atlas}, and studies geometric guardrails~\cite{aneja2026intrinsic}.

Four questions remain open across heterogeneous model families: (i) how folded Orthogonal Procrustes compares against full multi-layer~\cite{bakker2026relative} and fixed-depth Relative Representations, external text classifiers, and few-shot native probes across input-property, behavioral, and security regimes; (ii) when and why zero-hyperparameter mean-shift probes outperform source-tuned logistic regression; (iii) why flexible maps and whitening fit anchors better than rigid rotations yet trail in probe transfer (Figure~\ref{fig:map_side}), and how anchor domain diversity or unlabeled target-domain anchors affect robustness; and (iv) how much remaining transfer error multi-source averaging closes without target labels.

Across six models spanning four families (Qwen, Gemma, Meta Llama~3.1, and AllenAI OLMo~3), we make four contributions:
\begin{enumerate}
    \item \textit{Empirical boundaries and baselines:} We benchmark folded Orthogonal Procrustes against full multi-layer and fixed-depth Relative Representations, external text classifiers, and $k$-shot native probes across input-property, behavioral sycophancy, long-context, and adaptive security regimes (Section~\ref{sec:readonly}).
    \item \textit{Probe-side behavior and calibration:} Through a falsified anchor-projection test, shrinkage-LDA sweeps, eigenvector alignment, split-half disattenuation, and source-size tests, we show that mean-shift probes transfer best across core tasks for task-dependent reasons and verify held-out in-domain benign calibration (Section~\ref{sec:probe_side}).
    \item \textit{Map-side dissociation mechanism and anchor banks:} Extending Cheng et al.~\cite{cheng2026platoskey}, we show that whitened, Ridge, Kernel Ridge, and neural maps beat Scaled Procrustes in anchor fit yet trail in transfer because anisotropic shrinkage distorts target directions, and compare balanced four-domain against unlabeled in-domain anchor banks (Figure~\ref{fig:map_side}, Section~\ref{sec:map_side}).
    \item \textit{Four-family replication, two-way sensitivity, and multi-source averaging:} We test probe-side claims on new out-of-family pairs, audit pair non-independence with two-way cluster bootstrap CIs, evaluate leave-one-family-out multi-source averaging alongside a post-hoc combined recipe, and distill practical deployment rules (Sections~\ref{sec:four_family}--\ref{sec:limitations_guidance}).
\end{enumerate}

\section{Problem Formulation and Alignment Geometry}
\label{sec:method}

\begin{figure*}[t]
    \centering
    \includegraphics[width=0.69\textwidth]{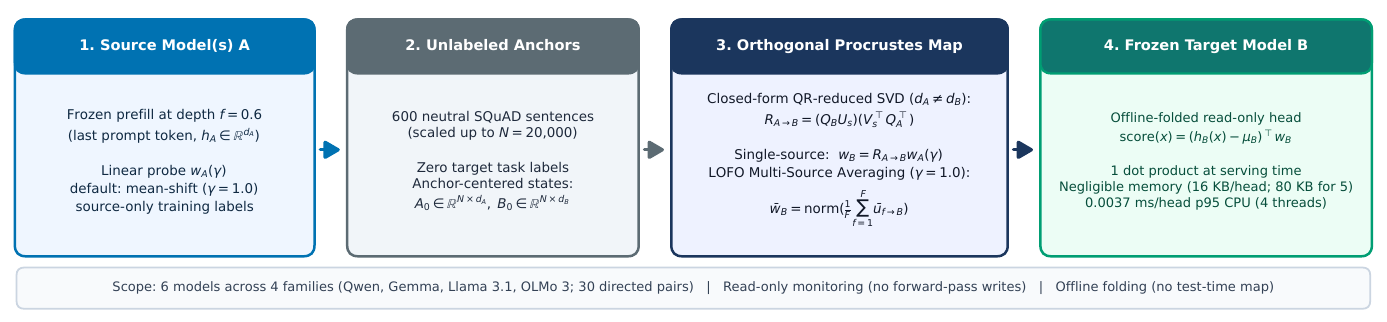}
    \caption{Read-only cross-model probe transfer folds offline into one target-space dot product with negligible memory (16\,KB/head). Source probes trained on frozen prefill states travel onto a frozen target model through an Orthogonal Procrustes rotation (or leave-one-family-out multi-source average) fitted on 600 unlabeled \texttt{SQuAD} anchors.}
    \label{fig:pipeline}
\end{figure*}

\paragraph{Models, Depths, and Non-Independence.}
Our initial benchmark evaluates 12 directed pairs across four dense and hybrid models: Qwen3-4B (36 layers, $d=2560$)~\cite{qwen2025qwen3}, Qwen3.5-9B (32-layer hybrid Gated DeltaNet/Attention, $d=4096$)~\cite{qwen2026qwen35,yang2025gateddeltanet}, Gemma-4-E4B-it (42 layers, $d=2560$)~\cite{gemma2026gemma4}, and Gemma-4-12B-it (48 layers, $d=3840$)~\cite{gemma2026gemma4}. Adding Meta Llama~3.1~8B Instruct~\cite{grattafiori2024llama3} (\texttt{NousResearch/Meta-Llama-3.1-8B-Instruct}) and AllenAI OLMo~3~7B Instruct~\cite{olmo2025olmo3} (32 layers, $d=4096$ each) yields 30 directed pairs including 18 new out-of-family pairs.

For every prompt, we apply the native chat template (thinking disabled, generation prompt enabled, pinned date \texttt{26 Jul 2024} on Llama~3.1 and OLMo~3) and extract the last-token prefill state $h_M^{(\ell)}(x) \in \mathbb{R}^{d_M}$ at locked depth $\ell(f) = \operatorname{round}(f \cdot L) - 1$. We fix $f=0.6$ primarily (secondary $f=0.5$ in Appendix~\ref{app:depth05}), avoiding target layer tuning and shallow CKA peaks~\cite{kornblith2019cka}. Because pairs share models, we report All-30 pair-medians with Holm--Bonferroni corrections, pair-cluster and 2,000-resample two-way (source $\times$ target) cluster bootstrap 95\% CIs, and leave-one-model-out / leave-one-family-out checks (Table~\ref{tab:app_v7_a5}). All pre-registered hypotheses and decision rules were recorded before execution and are archived in Appendix~\ref{app:prereg}.

\paragraph{Orthogonal Procrustes, Chaining, and Multi-Source Averaging.}
Given $N$ unlabeled neutral \texttt{rajpurkar/squad}~\cite{rajpurkar2016squad} anchor sentences ($N=600$ baseline, scaled up to $20{,}000$ in Section~\ref{sec:map_side}), let $A_0 \in \mathbb{R}^{N \times d_A}$ and $B_0 \in \mathbb{R}^{N \times d_B}$ denote anchor-centered activation matrices ($\mu_A, \mu_B$). When $N < \min(d_A, d_B)$ ($d_A \neq d_B$), reduced QR factorizations $A_0^\top = Q_A R_A$, $B_0^\top = Q_B R_B$ and an $N \times N$ SVD $R_B R_A^\top = U_s \Sigma V_s^\top$ (or direct SVD of $B_0^\top A_0$ for large $N$) yield the semi-orthogonal Procrustes map~\cite{schonemann1966procrustes}:
\begin{equation}
    R_{A \to B} = (Q_B U_{s,[:,:r]})(V_{s,[:,:r]}^\top Q_A^\top),
    \label{eq:op}
\end{equation}
retaining singular values $\sigma_i > 10^{-5}\sigma_1$ as a read-only stitch~\cite{lenc2015understanding,bansal2021stitching} (Figure~\ref{fig:pipeline}). Centered 600-anchor maps have rank at most 599 and preserve Euclidean inner products strictly inside that subspace.

We also evaluate two extensions. Two-hop \textit{disjoint-anchor chaining} ($A \to B \to C$) composes rotations fitted on non-overlapping 600-sentence pools. Leave-one-family-out \textit{multi-source averaging} unit-normalizes rotated mean-shift probes $u_{a \to B} = \operatorname{norm}(R_{a \to B} w_a(1.0))$ and averages them across out-of-family sources (family-balanced $\bar{w}_B = \operatorname{norm}(\frac{1}{F}\sum_f \bar{u}_{f \to B})$ or unweighted, agreeing within 0.002 AUROC). Offline folding reduces serving to one dot product (16\,KB/head in \texttt{fp32}; 80\,KB for 5 heads at $d_B=4096$, 0\,MB extra VRAM, 0.0037\,ms/head 4-thread CPU p95 vs.\ 158\,ms / 352\,MB for \texttt{deberta-v3-base-injection}~\cite{deepset_deberta_injection} and 42\,ms / 224\,MB for \texttt{MiniLM-L12-v2}~\cite{reimers2020multilingual}; Table~\ref{tab:app_security_ops}).

\paragraph{Probes, Tasks, and Centering.}
Following regularized discriminant analysis~\cite{friedman1989regularized} and covariance shrinkage~\cite{ledoit2004wellconditioned}, we interpolate continuously between covariance-inverted discriminant probes and mean-shift probes via scale-normalized shrinkage-LDA:
\begin{equation}
    \Sigma_W(\gamma) = (1 - \gamma)\Sigma_W + \gamma \frac{\mathrm{tr}(\Sigma_W)}{d} I, \quad w(\gamma) = \Sigma_W(\gamma)^{-1} \Delta\mu,
    \label{eq:shrink_lda}
\end{equation}
where $\Sigma_W = \frac{1}{n-2}\sum_{c \in \{0,1\}} (X_c - \mu_c)^\top(X_c - \mu_c)$, $\Delta\mu = \mu_1 - \mu_0$, and $w(\gamma)$ is unit-normalized across $\gamma \in \{0.001, \dots, 1.0\}$. Setting $\gamma=1.0$ recovers the zero-hyperparameter mean-shift (difference-of-means) direction, whereas $L_2$-logistic regression approximates $\Sigma_W^{-1}$ implicitly through its margin Hessian. On balanced, centered data, $L_2$-logistic regression converges to the mean-shift direction as $C \to 0$; stronger regularization therefore suppresses low-variance source directions that source-model cross-validation tends to under-regularize. We evaluate both fixed regularization ($C=1.0$ in initial Table~\ref{tab:readonly_main} comparisons) and 5-fold source cross-validated regularization ($C^*$, grouped by question ID on \texttt{TruthfulQA}). Because \texttt{liblinear} penalizes the intercept, training-mean versus anchor-mean centering can shift logistic weights when task and anchor means diverge, whereas shrinkage-LDA and mean-shift probes are shift-invariant (exact centering protocols appear in every table caption and Appendix~\ref{app:hyperparams}).

We group evaluations into three label regimes: \textit{input-property} tasks (\texttt{xTRam1}~\cite{xtram1_safeguard}, \texttt{JBB-Behaviors}~\cite{chao2024jailbreakbench}, length-matched \texttt{SQL}~\cite{bmc2_sql_create_context}, out-of-distribution \texttt{deepset}~\cite{deepset_prompt_injections}, length-matched \texttt{Anthropic/hh-rlhf}~\cite{bai2022hhrlhf}, \texttt{Paul/XSTest}~\cite{rottger2024xstest}, and length-matched \texttt{MATH-500}~\cite{hendrycks2021math,lightman2024verify}; hereafter \texttt{hh-rlhf} and \texttt{XSTest}), statement truthfulness (\texttt{TruthfulQA}~\cite{lin2022truthfulqa}), and template-matched Behavioral Sycophancy (derived from~\cite{perez2023discovering}), a \textit{model-behavior} task where label~1 indicates that the target model flips its greedy answer to match user bias. Early length- and template-shortcut splits were retired prior to main evaluation; \texttt{TruthfulQA} splits are question-ID-grouped, sycophancy splits are claim-disjoint, and the \texttt{xTRam1} training set for \texttt{deepset} was deduplicated to $n=392$ with verified zero exact or near-duplicate overlap against test and calibration splits (Appendix~\ref{app:hyperparams}). \codeRepoNote

\section{Read-Only Transfer Boundaries and Baselines}
\label{sec:readonly}

\begin{table*}[t]
\centering
\caption{Full OP beats Ridge and fixed-depth RelRep on OOD and cross-lingual injection, while two-hop disjoint chaining retains 95.7\%--99.9\% median above-chance AUROC (depth $0.6\to 0.6$, initial 12 pairs, 3-seed medians, $N=600$ \texttt{SQuAD}). Source \texttt{l2\_lr} uses fixed $C=1.0$ ($\mu_{\mathrm{tr}}$) here vs.\ source-CV $C^*$ ($\mu_{\mathrm{anc}}$) in Table~\ref{tab:cov_sweep} (\texttt{TruthfulQA} 0.733 vs.\ 0.802); transfer uses $\mu_{\mathrm{anc}}$ (\texttt{mean\_shift} is shift-invariant). Unified canonical All-30 pair-medians on deduplicated \texttt{xTRam1} ($n=392$)$\to$\texttt{deepset} ($n=116$, zero overlap) reach 0.835 (\texttt{l2\_lr}($C^*$), $\mu_{\mathrm{tr}}$) and 0.781 (\texttt{mean\_shift}; Tables~\ref{tab:v8_bakker}, \ref{tab:v8_mixed}, and~\ref{tab:app_deepset_v6}).}
\label{tab:readonly_main}
\footnotesize
\setlength{\tabcolsep}{3.4pt}
\begin{tabular}{@{}lccccccc@{}}
\toprule
\textbf{Task} & \textbf{Native} & \textbf{Full OP} & \textbf{Ridge CV} & \textbf{RelRep Raw} & \textbf{RelRep Cen.} & \textbf{Full OP (\texttt{ms})} & \textbf{Disjoint 24-Chain} \\
 & \textbf{(Med 4)} & \textbf{(\texttt{l2\_lr}, $C{=}1$, $\mu_{\mathrm{tr}}$)} & \textbf{(\texttt{l2\_lr}, $C{=}1$, $\mu_{\mathrm{tr}}$)} & \textbf{(Moschella)} & \textbf{(Centered)} & \textbf{(Diag.)} & \textbf{Ret.\ Med [Min]} \\
\midrule
xTRam1 Prompt Inj. & 0.9994 & 0.9902$\pm$0.0011 & 0.9861 & 0.9357 & 0.9654 & 0.9814 & 99.91\% [96.41\%] \\
JBB-Behaviors & 0.9862 & 0.9489$\pm$0.0078 & 0.9187 & 0.9288 & 0.9174 & 0.9606 & 97.24\% [47.10\%] \\
SQL vs.\ SQuAD (Len) & 0.9980 & 0.9890$\pm$0.0004 & 0.9914 & 0.9733 & 0.9878 & 0.9885 & 99.78\% [95.40\%] \\
xTRam1 ($n=392$)$\to$deepset & 0.9496 & 0.8455$\pm$0.0116 & 0.8100 & 0.4326 & 0.6084 & 0.7832 & --- \\
Turkish PI (Gemma-3-transl.) & 0.9820 & 0.9604$\pm$0.0048 & 0.9537 & 0.8724 & 0.9133 & 0.9564 & --- \\
TruthfulQA (QID) & 0.9095 & 0.7329$\pm$0.0091 & 0.7262 & 0.7125 & 0.7459 & 0.7851 & 95.69\% [63.19\%] \\
MATH-500 ($\pm 15\%$ Len) & 0.8150 & 0.7440$\pm$0.0290 & 0.7324 & 0.7988 & 0.7771 & 0.8048 & 95.69\% [47.75\%] \\
\bottomrule
\end{tabular}
\end{table*}

\begin{figure*}[t]
    \centering
    \includegraphics[width=0.73\textwidth]{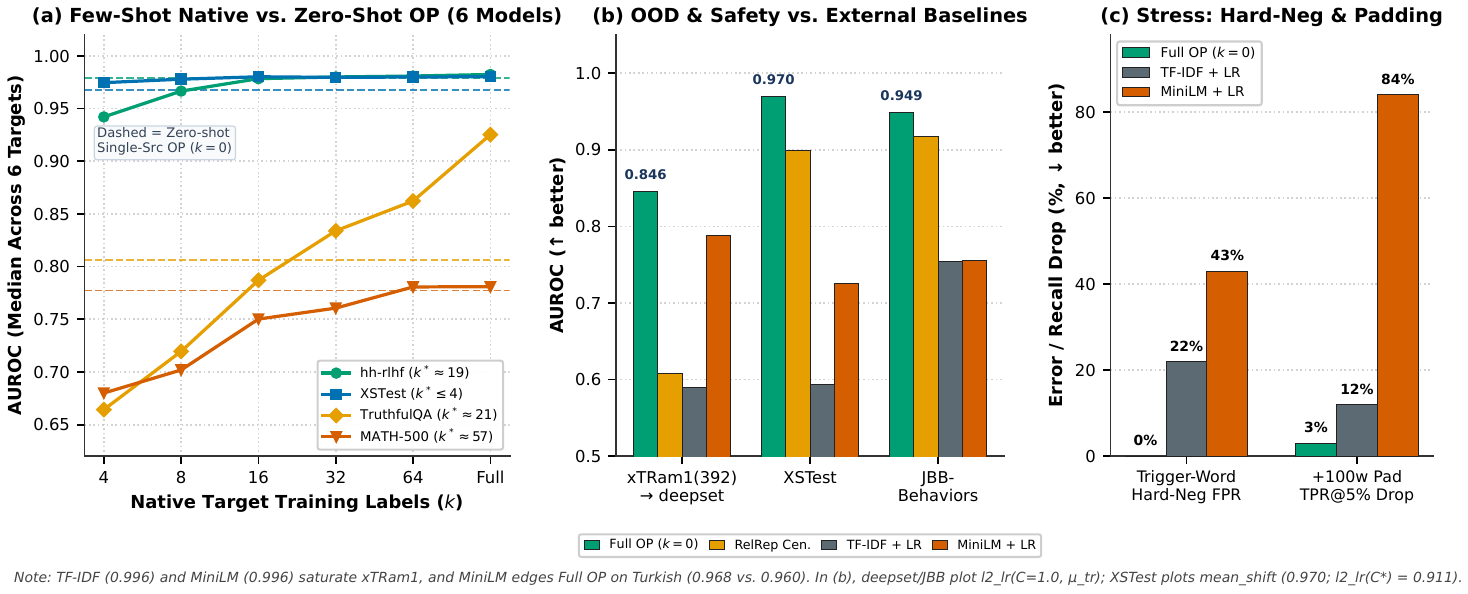}
    \caption{Zero-shot Orthogonal Procrustes beats external text baselines under semantic and lexical shifts and matches $k^* \le 4$ to $57$ native target labels (depth $0.6\to 0.6$). (a) Canonical 6-model single-source mean-shift transfer ($k=0$ dashed lines, $N=600$) vs.\ best $k$-shot native target probes (multi-source $k^*$ and partial-access prior-regularized probes in Table~\ref{tab:app_v7_a4}). (b) Initial 12-pair OOD and safety AUROC (\texttt{xTRam1} $n=392$ $\to$ \texttt{deepset} $n=116$ and \texttt{JBB} use source-centered logistic probes at $C=1.0$; \texttt{XSTest} uses mean-shift) vs.\ fixed-depth RelRep, TF-IDF, and MiniLM. (c) Trigger-word hard-negative FPR ($\tau_{95}$) and $+100$-word benign-padding recall drop (TPR@5\%FPR).}
    \label{fig:sample_eff}
\end{figure*}

\subsection{Alignment Methods and Multi-Layer Relative Representations}
\label{sec:baselines_align}

Across the initial 12 model pairs with 600 \texttt{SQuAD} anchors, Orthogonal Procrustes outperforms Ridge regression and fixed-depth cosine Relative Representations~\cite{moschella2023relative} whenever the test distribution shifts away from training phrasing (Table~\ref{tab:readonly_main}). On deduplicated \texttt{xTRam1} ($n=392$) $\to$ \texttt{deepset} ($n=116$), source-centered logistic probes at $C=1.0$ reach 0.85 AUROC under Procrustes versus 0.81 under Ridge and 0.61 under fixed-depth centered Relative Representations (full All-30 centering audit in Table~\ref{tab:app_deepset_v6}). Raw uncentered Relative Representations drop to 0.43 due to anchor-similarity shift and sign inversion across families when task and anchor means diverge. On \texttt{XSTest} (trained on \texttt{JBB}), Procrustes gains $+52$\,pp TPR@1\%FPR with mean-shift probes over fixed-depth centered Relative Representations, widening across all four families (Section~\ref{sec:four_family}).

Folded depth-matched Procrustes also outperforms full multi-layer global Relative Representations~\cite{bakker2026relative} by a wide margin at a fraction of the serving cost (Table~\ref{tab:v8_bakker}). Native multi-layer Bakker and Louizos probes reach 0.75--0.98 AUROC across tasks, confirming implementation fidelity. Across all 30 pairs from our unified canonical extraction, folded depth-0.6 Procrustes beats Bakker's logistic configuration (raw global concatenation, 0.712 four-task median) by $+0.108$ AUROC with matched source-CV logistic probes (50.7\% vs.\ 6.2\% \texttt{XSTest} recall) and $+0.158$ with mean-shift probes ($+0.190$ and 81.1\% vs.\ 40.6\% recall against matched Bakker mean-shift; all margins hold strictly positive two-way 95\% CIs in Table~\ref{tab:app_v7_a5}).

Per-layer $L_2$ normalization lifts Bakker transfer to 0.735, showing that cross-layer norm differences across architectures hurt raw concatenation, yet Procrustes mean-shift still leads by $+0.147$. At serving time on 4 CPU threads, Bakker global hooks all 32--48 layers and stores 148.2\,MB of \texttt{fp16} anchors at 6.94\,ms p95 latency, versus one layer, 7\,KB, and 0.0012\,ms for folded Procrustes (Table~\ref{tab:v8_bakker}).

\subsection{External Text Classifiers and Few-Label Target Probes}
\label{sec:baselines_ext}

Zero-shot Procrustes transfer outperforms external text classifiers under semantic and lexical shifts (Figure~\ref{fig:sample_eff}b--c). While TF-IDF and multilingual MiniLM saturate in-distribution \texttt{xTRam1} (0.996) and MiniLM edges Procrustes on Turkish prompt injection (0.97 vs.\ 0.96), both degrade under shift: on \texttt{XSTest}, TF-IDF (0.59) and MiniLM (0.73) trail transferred mean-shift probes (0.97); on \texttt{JBB}, both score 0.76 vs.\ 0.95; and on \texttt{xTRam1} ($n=392$) $\to$ \texttt{deepset}, TF-IDF (0.59) and MiniLM (0.79) trail source-centered Procrustes logistic probes (0.85). On 100 author-audited benign trigger-word prompts, TF-IDF and MiniLM suffer 22\% and 43\% false-alarm rates versus 0\% for Procrustes, and adding 100 benign padding words drops MiniLM recall by 84\,pp versus 3\,pp for Procrustes.

When source training texts remain accessible, retraining natively on the target model is inexpensive and usually preferable; cross-model transfer targets deleted, private, or vendor-locked corpora. Across all six target models against the best few-label native probe (Figure~\ref{fig:sample_eff}a), single-source mean-shift transfer with 600 anchors matches $k^* \le 4$ target labels on \texttt{XSTest}, 19 on \texttt{hh-rlhf}, 21 on \texttt{TruthfulQA}, and 57 on \texttt{MATH-500}, while multi-source averaging raises equivalence to 6 on \texttt{XSTest}, 27 on \texttt{TruthfulQA}, and over 64 on \texttt{hh-rlhf} and \texttt{MATH-500}. In a paired partial-access check (Table~\ref{tab:app_v7_a4}), prior-regularized target probes anchored to the transferred direction met our pre-registered bar on just one of four tasks (\texttt{TruthfulQA} at $k=16$), so we do not claim general partial-access gains.

\subsection{Behavioral Labels, Scope, and Security Boundaries}
\label{sec:scope_security}

Transfer succeeds when labels depend on the input prompt, yet degrades when labels depend on target-model behavior. On input-property tasks, transferred mean-shift probes retain 94\%--99.5\% of matched native above-chance AUROC (\texttt{hh-rlhf}: 0.98/0.98; \texttt{XSTest}: 0.97/0.98; \texttt{MATH-500}: 0.78/0.79), and on \texttt{TruthfulQA} they reach 0.82 (86\% of native mean-shift, 74\% of native logistic). On Behavioral Sycophancy, by contrast, transfer drops from 0.82 native to 0.62 transferred for logistic probes and from 0.78 to 0.71 for mean-shift probes (Table~\ref{tab:app_v7_a3}; archived-run parity in Appendix~\ref{app:hyperparams}). Our three-term decomposition\footnote{On symmetric all-pairs slices, every model appears equally often as source and target, so the average native predictability difference between target and source cancels to zero.} shows that on All-300 prompts, mean-shift's drop splits 31\% label shift and 69\% direction loss, whereas on shared counter-bias prompts 98.3\% reflects cross-model label disagreement ($\kappa=0.26$) with near-zero direction loss (logistic probes split evenly across both terms; Table~\ref{tab:app_v7_a3}).

Four additional scope and security checks bound deployment (Tables~\ref{tab:readonly_main} and~\ref{tab:app_security_ops}, Appendix~\ref{app:hyperparams}). First, two-hop disjoint chaining retains 95.7\%--99.9\% median direct above-chance AUROC across five tasks, though worst chains drop to 47.1\%. Second, held-out Turkish prompt injection (Gemma-3-translated~\cite{gemma2025gemma3}) retains 98\% of same-model cross-lingual transfer (0.96 vs.\ 0.98; 12/12 Holm $p<0.05$). Third, on 300 4K-token indirect-injection documents, short-trained probes drop to 0.72--0.76 AUROC (similar to native short-to-4K transfer at 0.77, below 4K TF-IDF at 0.84; cause unestablished). Fourth, under 20-query adaptive suffix evasion, single-layer Procrustes drops from 0.99 to 0.95 AUROC (50\% TPR@1\%FPR), whereas a zero-VRAM four-layer ensemble cuts AUROC loss by 35\% and raises TPR@1\%FPR to 64\% (native: 83\%).

\section{Probe-Side Geometry: Covariance Inversion and Calibration}
\label{sec:probe_side}

Under cross-model transfer, simple mean-shift probes consistently outperformed logistic probes on \texttt{XSTest}, \texttt{MATH-500}, and Behavioral Sycophancy, edged them on \texttt{TruthfulQA}, and tied on \texttt{hh-rlhf}. To understand why, we tested pre-registered hypotheses (rules archived in Appendix~\ref{app:prereg}) alongside targeted diagnostic checks.

\subsection{Falsified Anchor-Subspace Projection}
\label{sec:svop}

We first hypothesized that logistic probes degrade under transfer by placing weight outside the leading 600-anchor singular subspace. That hypothesis failed cleanly: subspace energy showed zero positive correlation with retention ($\rho=+0.01$ across the four core tasks, $-0.22$ across all seven evaluation tasks), and projecting probes into the anchor subspace (SV-OP) helped \texttt{TruthfulQA} ($+0.066$) yet hurt \texttt{XSTest} ($-0.091$) and \texttt{MATH-500} ($-0.121$). We discarded SV-OP and tested covariance inversion directly.

\subsection{Covariance-Inversion Sweep and Source-CV Pitfall}
\label{sec:cov_sweep}

\begin{figure*}[t]
    \centering
    \includegraphics[width=0.59\textwidth]{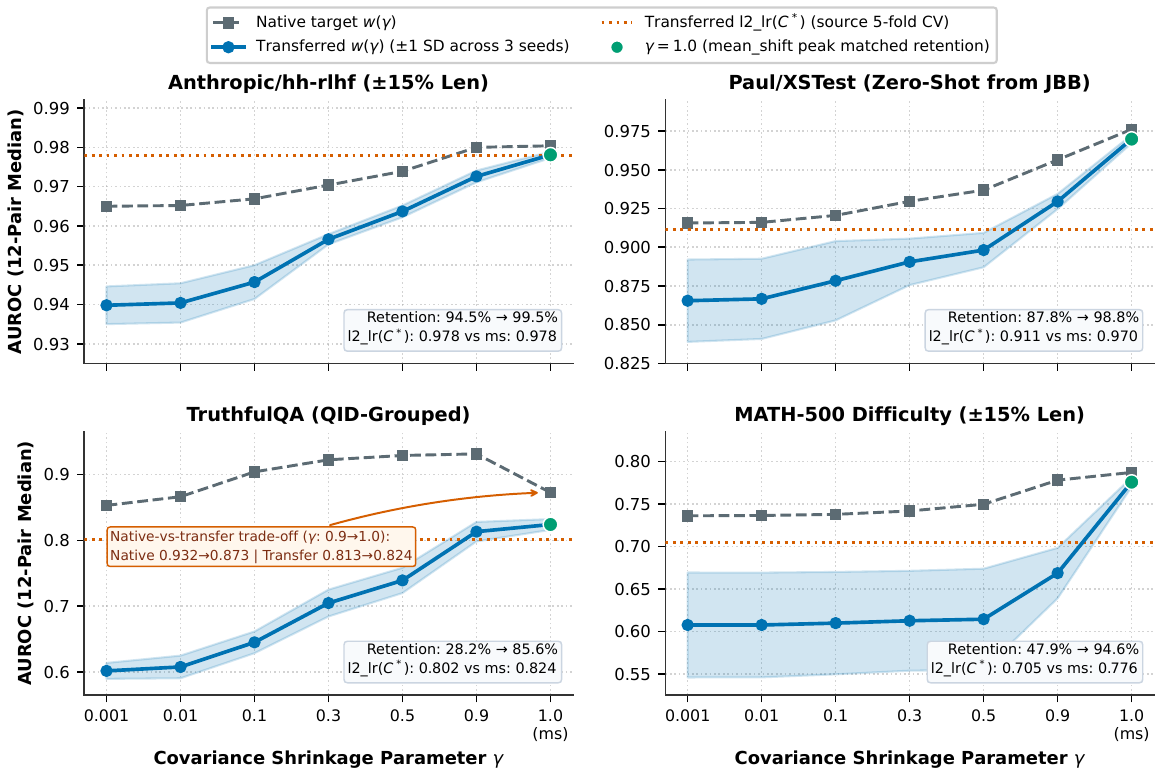}
    \caption{Shrinking $\Sigma_W(\gamma)$ toward the identity monotonically raises transfer AUROC on all four core tasks and matches or beats source-CV logistic probes (dotted lines; depth $0.6\to 0.6$, initial 12 pairs, $\pm 1$\,SD across 3 seeds). On \texttt{TruthfulQA} ($\gamma: 0.9 \to 1.0$), native AUROC drops ($0.932 \to 0.873$) while transfer rises ($0.813 \to 0.824$), exposing a native-versus-transfer trade-off.}
    \label{fig:cov_sweep}
\end{figure*}

\begin{table*}[t]
\centering
\caption{Shrinking $\Sigma_W(\gamma)$ monotonically raises transfer AUROC and peaks in matched native above-chance retention at $\gamma=1.0$ (\texttt{mean\_shift}, shift-invariant) on all 4 core tasks, matching or beating source-CV \texttt{l2\_lr}($C^*$) (depth $0.6\to 0.6$, initial 12 pairs, 3 seeds; default centering, Section~\ref{sec:method}).}
\label{tab:cov_sweep}
\footnotesize
\setlength{\tabcolsep}{3.5pt}
\begin{tabular}{@{}lccccccc@{}}
\toprule
\textbf{Task} & \textbf{$\gamma=0.001$} & \textbf{$\gamma=0.1$} & \textbf{$\gamma=0.5$} & \textbf{$\gamma=0.9$} & \textbf{$\gamma=1.0$ (\texttt{ms})} & \textbf{Matched Ret.} & \textbf{Source CV \texttt{l2\_lr}($C^*$)} \\
 & \textbf{Trans / Nat} & \textbf{Trans / Nat} & \textbf{Trans / Nat} & \textbf{Trans / Nat} & \textbf{Trans / Nat} & \textbf{$\gamma:0.001\to 1.0$} & \textbf{Trans ($\cos(w,\Delta\mu)$)} \\
\midrule
hh-rlhf & 0.940 / 0.965 & 0.946 / 0.967 & 0.964 / 0.974 & 0.973 / 0.980 & 0.978 / 0.980 & 94.5\%$\to$99.5\% & 0.9779 (0.48) \\
XSTest (JBB) & 0.866 / 0.916 & 0.878 / 0.921 & 0.898 / 0.937 & 0.930 / 0.957 & 0.970 / 0.976 & 87.8\%$\to$98.8\% & 0.9114 (0.34) \\
TruthfulQA (QID) & 0.602 / 0.853 & 0.645 / 0.904 & 0.739 / 0.929 & 0.813 / 0.931 & 0.824 / 0.873 & 28.2\%$\to$85.6\% & 0.8018 (0.45) \\
MATH-500 ($\pm 15\%$) & 0.608 / 0.736 & 0.610 / 0.738 & 0.614 / 0.750 & 0.669 / 0.778 & 0.776 / 0.787 & 47.9\%$\to$94.6\% & 0.7048 (0.52) \\
\bottomrule
\end{tabular}
\end{table*}

Shrinking within-class covariance toward the identity via~\eqref{eq:shrink_lda} consistently improves cross-model transfer. Weakly regularized logistic probes and unshrunk discriminant probes amplify low-variance source directions, echoing the noise sensitivity of whitened inner products noted by Svendsen et al.~\cite{svendsen2026improving}. Across the initial 12 pairs at depth 0.6, increasing shrinkage $\gamma \to 1.0$ raised transfer AUROC monotonically on all four core tasks, decreasing logistic $C$ did so on three of four tasks, and matched retention peaked at mean-shift ($\gamma=1.0$) on all four tasks (86\%--99.5\%; Figure~\ref{fig:cov_sweep}, Table~\ref{tab:cov_sweep}; depth 0.5 in Appendix~\ref{app:depth05}).

On \texttt{TruthfulQA}, moving $\gamma$ from 0.9 to 1.0 lowers native AUROC (0.93 to 0.87) while raising transfer AUROC (0.81 to 0.82), exposing a native-versus-transfer trade-off that aligns with concurrent work by Sadeghi and Becker~\cite{sadeghi2026probing}. Because $L_2$-logistic regression approaches the mean-shift direction as $C \to 0$ on balanced, centered data, selecting $C^*$ by source cross-validation for native accuracy systematically under-regularizes for transfer, causing source-CV logistic probes to trail mean-shift on \texttt{XSTest} (0.91 vs.\ 0.97), \texttt{MATH-500} (0.70 vs.\ 0.78), and \texttt{TruthfulQA} (0.80 vs.\ 0.82; Table~\ref{tab:cov_sweep}).

\subsection{Safe-Prompt False-Alarm Calibration on XSTest}
\label{sec:xstest_cal}

Threshold calibration must use in-domain benign traffic rather than out-of-distribution text, and methods should be compared by unsafe recall at a fixed false-alarm rate. Calibrating a 5\% threshold on benign \texttt{xTRam1} prompts (\texttt{cal200}) inflates \texttt{XSTest} safe-prompt false alarms (probe alarms, not refusals) to 31\% for transferred mean-shift and 27\% for native mean-shift alike (Figure~\ref{fig:xstest_cal}). At matched 50\%--90\% unsafe recall, transferred mean-shift holds 0.2\%--9.8\% false alarms, well below source-CV logistic probes and fixed-depth centered Relative Representations (Figure~\ref{fig:xstest_cal}a).

Under held-out in-domain benign calibration (100 random 50/50 splits of 250 safe prompts at a 5\% target across all 30 pairs; Table~\ref{tab:app_v7_a2}), landing near 5.0\% held-out false-alarm rate holds by exchangeability for any score function (meeting our $\le 8\%$ pre-registered check). The discriminative separation lies in unsafe recall at that calibrated false-alarm rate: transferred mean-shift reaches 81.1\% unsafe recall versus 50.7\% for source-CV logistic probes, 17.7\% for anchor-centered fixed-depth Relative Representations, 6.2\% for Bakker's logistic configuration (40.6\% for Bakker mean-shift; Table~\ref{tab:v8_bakker}), and 86.0\% for native mean-shift.

\subsection{Eigenvector Alignment, Split-Half, and Sample-Size Diagnostics}
\label{sec:eig_decomp}

\begin{table*}[t]
\centering
\caption{Split-half disattenuation separates low-variance estimation noise, 10.5\%--30.4\% anchor-span loss, and persistent cross-model mismatch ($m(\gamma)<1$) at depth $0.6\to 0.6$ ($n=36$ Original-12 pair-seed entries; shift-invariant probes, default centering). Guarded $\Delta R^2_{\mathrm{cross}} = R^2_{\mathrm{cross}} - R^2_{\mathrm{rand}}$ is descriptive. $^*$Low-power ($n=54$/half).}
\label{tab:split_half}
\scriptsize
\setlength{\tabcolsep}{2.7pt}
\begin{tabular}{@{}lccccccc@{}}
\toprule
\textbf{Task} & \textbf{$\cos_{\mathrm{within}}$} & \textbf{Span Loss} & \textbf{$m(0.3)$ [95\% CI]} & \textbf{$m(1.0)$ [95\% CI]} & \textbf{Within Overlap} & \textbf{Guarded $\Delta R^2_{\mathrm{cross}}$} & \textbf{Pre-Registered} \\
 & \textbf{($\gamma:0.3\to 1.0$)} & \textbf{$1-\rho^*_{\mathrm{span}}(1.0)$} & \textbf{(Entry CI)} & \textbf{(Entry CI)} & \textbf{(Q1 / Q4)} & \textbf{(Q1 / Q4)} & \textbf{Classification} \\
\midrule
hh-rlhf (150/half) & 0.088$\to$0.974 & 10.5\% & 0.482 [0.419, 0.601] & 0.625 [0.609, 0.635] & 0.554 / 0.045 & +0.021 / +0.042 & Shrinkage reduces mismatch \\
XSTest (100/half) & 0.157$\to$0.974 & 15.9\% & 0.473 [0.329, 0.559] & 0.506 [0.466, 0.524] & 0.530 / 0.049 & +0.021 / +0.037 & Genuine mismatch persists \\
TruthfulQA (150/half) & 0.178$\to$0.948 & 18.3\% & 0.228 [0.171, 0.255] & 0.371 [0.336, 0.415] & 0.508 / 0.049 & +0.392 / +0.227 & Shrinkage reduces mismatch \\
MATH-500$^*$ (54/half) & 0.080$\to$0.781 & 30.4\% & -0.009 [-0.065, 0.209] & 0.209 [0.186, 0.229] & 0.544 / 0.029 & +0.044 / +0.056 & Genuine mismatch persists \\
\bottomrule
\end{tabular}
\end{table*}

High-variance source directions align far better across models than low-variance directions. Measuring held-out Pearson correlation $r_k$ across source within-class eigenvectors $v_k$ met all pre-registered gates on 48/48 initial pair-task entries (Figure~\ref{fig:mechanism}a, Table~\ref{tab:app_pairs_eig}): Spearman correlation between variance and alignment sits at $+0.44$ to $+0.59$, and cross-model probe cosine rises monotonically with $\gamma$ on all four tasks (though high AUROC does not require high cosine: \texttt{MATH-500} reaches 0.13 cosine yet 95\% retention).

To separate finite-sample noise, anchor-span loss, and genuine cross-model mismatch, we split training sets into disjoint halves and apply Spearman disattenuation~\cite{spearman1904proof}:
\begin{equation}
    \rho^*(\gamma) = \frac{\tfrac{1}{2}[\cos(R w_A^{(1)}, w_B^{(2)}) + \cos(R w_A^{(2)}, w_B^{(1)})]}{\sqrt{\cos_{\mathrm{within},A}(\gamma)\,\cos_{\mathrm{within},B}(\gamma)}},
    \label{eq:disatt}
\end{equation}
normalizing by the anchor-span ceiling $\rho^*_{\mathrm{span}}(\gamma)$ to obtain $m(\gamma) = \rho^*(\gamma)/\rho^*_{\mathrm{span}}(\gamma)$ (Table~\ref{tab:split_half}). Three factors appear: severe finite-sample instability in low-variance directions (within-model reliability rises from 0.07--0.18 at $\gamma=0.3$ to 0.78--0.97 at $\gamma=1.0$), 10.5\%--30.4\% anchor-span loss at 600 anchors, and genuine cross-model mismatch already present in leading directions ($m(\gamma) < 1$ throughout).

Finally, a pre-registered sample-size test predicted that the mean-shift advantage over source-CV logistic regression would shrink as source training size grows (Table~\ref{tab:app_check5_ncurve}). Zero of four tasks met this criterion, and the \texttt{XSTest} gap widened ($+0.034 \to +0.064$). Two diagnostic checks explain why (Table~\ref{tab:app_exploratory_n}): on \texttt{TruthfulQA} at fixed $\gamma=0.1$, larger source sets improve native AUROC (0.83 to 0.89) while hurting transfer (0.72 to 0.62) as probes exploit source-specific directions; on native \texttt{JBB} $\to$ \texttt{XSTest}, larger training sets cause source CV to weaken regularization, hurting logistic probes (0.98 to 0.96) while mean-shift remains flat.

\section{Map-Side Geometry: Better Anchor Fit Does Not Improve Transfer}
\label{sec:map_side}

\begin{figure*}[t]
    \centering
    \includegraphics[width=0.85\textwidth]{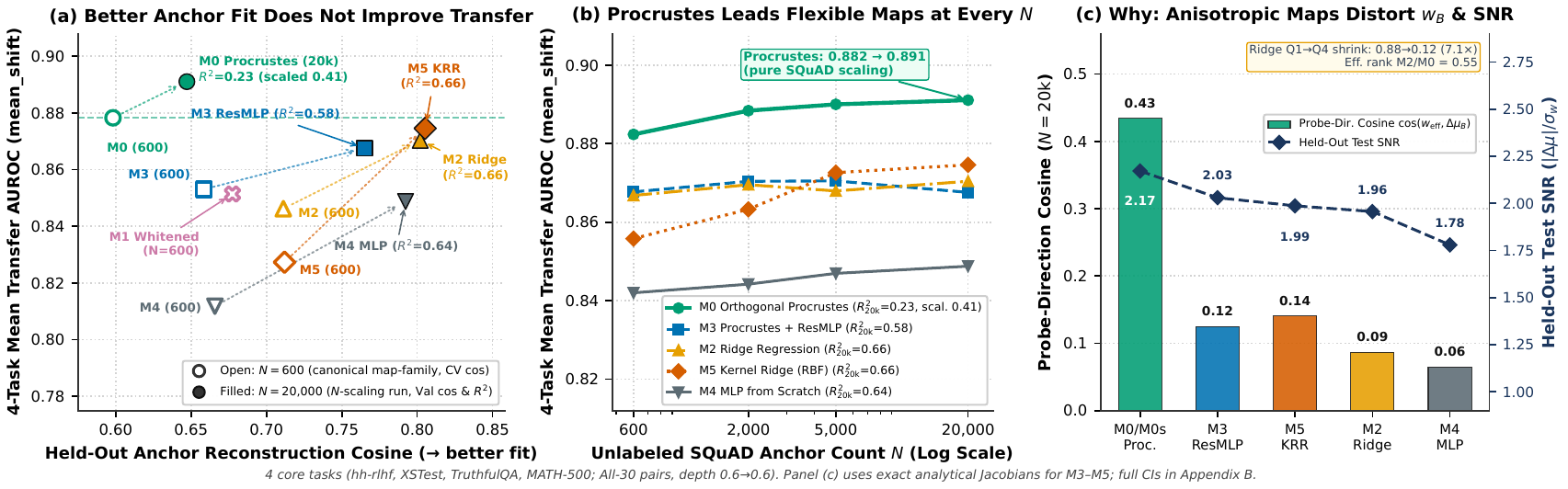}
    \caption{Every non-orthogonal map fits anchors better yet trails Orthogonal Procrustes in transfer because anisotropic shrinkage distorts target discriminative directions (all 30 pairs, 4 core tasks, depth $0.6\to 0.6$, mean-shift). (a) Anchor cosine (canonical $N=600$ CV cos; Val $R^2$ at $N=20{,}000$, Scaled Procrustes $R^2=0.41$) vs.\ 4-task AUROC across \texttt{M0}--\texttt{M5}. (b) Scaling \texttt{SQuAD} anchors $33\times$ ($N \in \{600..20\text{k}\}$). (c) At $N=20{,}000$, \texttt{M0}/\texttt{M0s} preserves $3.1\times$--$6.7\times$ higher probe cosine (4/4 tasks across \texttt{M2}--\texttt{M5}) and higher mean test SNR (4/4 CIs $>0$ vs.\ \texttt{M2}--\texttt{M4}, 2/4 vs.\ \texttt{M5} at $20\text{k}$, 4/4 at $600$) while Ridge compresses Q4 anchor modes $7.1\times$ harder than Q1.}
    \label{fig:map_side}
\end{figure*}

Because cross-model mismatch persists even in leading directions ($m(1.0) < 1$), we tested whether whitened or nonlinear alignment maps can close it, extending Cheng et al.'s~\cite{cheng2026platoskey} small-$N$ factual-swap observations to read-only probe transfer across all 30 model pairs up to 20,000 anchors. To score frozen source probes uniformly across \texttt{M0}--\texttt{M5}, every map is fitted target-to-source ($g_{B\to A}$) on anchor-centered, RMS-normalized states $\tilde{h}_B$ and scored as $w_A^\top g_{B\to A}(\tilde{h}_B)$ (exact for orthogonal maps, where $g_{B\to A} = R_{A\to B}^\top$). Linear maps (\texttt{M0}--\texttt{M2}) fold offline into one dot product, whereas nonlinear maps (\texttt{M3}--\texttt{M5}) require per-prompt hidden-layer or kernel evaluation.

\paragraph{Whitened Rotation and Scaling to 20,000 Anchors.}
At 600 anchors across all 30 pairs, replacing Orthogonal Procrustes (\texttt{M0}) with Whitened Rotation (\texttt{M1}) raised cross-validated anchor cosine (0.68 vs.\ 0.60) yet lowered mean-shift transfer across all four core tasks (0.85 vs.\ 0.88 four-task median; Figure~\ref{fig:map_side}a, Table~\ref{tab:app_v6_mapfamily}; depth 0.5 in Table~\ref{tab:app_v6_05}). Ridge (\texttt{M2}), Procrustes+ResMLP (\texttt{M3}), scratch MLP (\texttt{M4}), and RBF Kernel Ridge (\texttt{M5}) likewise achieved higher anchor cosine (0.66--0.71) yet trailed Procrustes on every task (0.81--0.85), with all five four-task margins keeping strictly positive two-way 95\% CIs (Table~\ref{tab:app_v7_a5}).

Scaling nested \texttt{SQuAD} anchors $33\times$ from 600 to 20,000 sentences rules out small-$N$ data starvation (Figure~\ref{fig:map_side}b, Table~\ref{tab:app_v6_scaling_4tasks}; \texttt{deepset} scaling in Table~\ref{tab:app_deepset_v6}). Optimal scalar Scaled Procrustes ($s^* R_{A\to B}$) leaves AUROC invariant while explaining 40.6\%--44.6\% of the anchor $R^2$ gap (Table~\ref{tab:app_v7_a1}). Yet Ridge, Kernel Ridge, and neural maps still beat Scaled Procrustes in validation $R^2$ (0.58--0.66 vs.\ 0.41) and cosine (0.77--0.81 vs.\ 0.65), while Orthogonal Procrustes leads at every anchor count on both mean-shift (0.882 to 0.891) and source-CV logistic probes (0.820 at $N=600$ in Table~\ref{tab:app_v6_mapfamily} to 0.863 at $N=20{,}000$ in Table~\ref{tab:app_v6_scaling_4tasks}; zero of four tasks met the nonlinear improvement bar) and folds offline (0.003\,ms vs.\ 0.98--20.5\,ms CPU p95).

\paragraph{Why Better Anchor Fit Fails to Transfer.}
Paralleling cross-lingual word-embedding alignment~\cite{xing2015normalized,conneau2018word}, orthogonality outperforms unconstrained linear and nonlinear maps across LLM residual streams; evaluating effective readout directions (linear pullbacks for \texttt{M0}--\texttt{M2}; analytical Jacobians for \texttt{M3}--\texttt{M5}) and held-out test signal-to-noise ratio reveals the underlying anisotropic-shrinkage and SNR mechanism (Figure~\ref{fig:map_side}c, Table~\ref{tab:v8_map_mech}). Although 97.3\% of 20,000-anchor \texttt{SQuAD} variance sits in top-quartile singular modes (Q1), 19.6\% of native \texttt{TruthfulQA} and 39.6\% of \texttt{MATH-500} probe energy sits in Q2--Q4. To minimize neutral-text reconstruction error, Ridge compresses Q4 modes $7.1\times$ harder than Q1 (effective rank ratio 0.55), whereas Orthogonal Procrustes preserves inner products isotropically.

As a result, scaling from 600 to 20,000 anchors lifts Procrustes probe-direction cosine against the native target direction from 0.284 to 0.434 (mean test SNR 2.05 to 2.17), while Ridge, ResMLP, MLP, and Kernel Ridge retain $3.1\times$--$6.7\times$ lower probe cosine (0.065--0.141) and lower mean test SNR (1.78--2.03; Table~\ref{tab:v8_map_mech}). Our pre-registered mechanism criterion was met against all four alternatives at both 600 and 20,000 anchors, with strictly positive pair-cluster 95\% CIs on 4/4 tasks in probe cosine across \texttt{M2}--\texttt{M5} and 4/4 in SNR against \texttt{M2}--\texttt{M4} (2/4 against \texttt{M5} at 20,000, 4/4 at 600).

\paragraph{Balanced Four-Domain and Unlabeled In-Domain Anchor Banks.}
While pure \texttt{SQuAD} scaling yields modest gains on the four core tasks (0.88 to 0.89), anchor domain composition strongly shapes transfer (Table~\ref{tab:v8_mixed}). Replacing 100\% \texttt{SQuAD} prose with a balanced four-domain bank (prose, code, multi-turn dialogue, and 10-language non-English text; zero test overlap) matches \texttt{SQuAD} on the four core tasks at 600 and 5,000 anchors while lifting \texttt{xTRam1} ($n=392$) $\to$ \texttt{deepset} at $N=600$ by $+0.095$ for mean-shift (0.781 to 0.887) and $+0.099$ for source-centered logistic probes (0.835 to 0.944, 99.1\% pair-median native above-chance AUROC; two-way CIs $>0$), beating 5,000 \texttt{SQuAD} anchors with $8.3\times$ fewer prompts. Single-domain ablations trace this gain to dialogue and multilingual coverage while showing that balanced mixing avoids the \texttt{TruthfulQA} drop ($-0.028$) caused by dialogue-only anchors.

With disjoint unlabeled target-domain anchors ($N \le 600$, ROUGE-L $<0.80$), Procrustes lifts four-task transfer to 0.902 ($+0.021$ over \texttt{SQuAD} and mixed, two-way CIs $>0$), beating \texttt{SQuAD} on 3/5 domains (\texttt{XSTest} $+0.012$, \texttt{TruthfulQA} $+0.047$ to 0.857, \texttt{deepset} $+0.136$ to 0.914/0.956) yet clearing two-way CIs over mixed anchors on 1/5 (\texttt{TruthfulQA} $+0.045$, where template matching contributes; \texttt{deepset} $+0.016$ crosses zero; Table~\ref{tab:v8_mixed}). On the same anchors, a 1D score-distillation baseline (target-space CV Ridge predicting scalar source scores $w_A^\top \tilde{h}_A(x_i)$, folding identically into one dot product) trails Procrustes on \texttt{SQuAD} ($+0.034$ four-task mean-shift, two-way CI $>0$) and mixed anchors ($+0.013$, two-way CI $>0$) where Procrustes shares one task-agnostic rotation; on unlabeled in-domain anchors it ties mean-shift (0.903 vs.\ 0.902) and slightly edges source-CV logistic probes (\texttt{deepset} 0.972 vs.\ 0.956) at the cost of per-probe refitting.

\section{Four-Family Replication and Multi-Source Averaging}
\label{sec:four_family}

\begin{table*}[t]
\centering
\caption{Across four families, \texttt{mean\_shift} (shift-invariant; default centering) retains 87\%--99\% native above-chance AUROC on New-18 out-of-family pairs while multi-source averaging closes 38\%--46\% of non-ceiling gaps (depth $0.6\to 0.6$, adding \texttt{Llama-3.1-8B} and \texttt{OLMo-3-7B}). Right: independent 6-target column medians ($N=600$; unweighted within 0.002: 0.9822/0.9772/0.8231/0.7816) and the post-hoc Combined Best Recipe$^{\ddagger}$ ($N=20\text{k}$ + multi-source); $\Delta$AUROC reports medians of paired $\Delta$ rather than differences of rounded column medians (\texttt{TruthfulQA} median $+0.018$, two-way OOF CI $[+0.003, +0.038]$ in Table~\ref{tab:app_v7_a5}). Gap-closed $(\text{Multi}-\text{Sing.})/(\text{Nat.}-\text{Sing.})$ takes the median over targets with $\text{Nat.}>\text{Sing.}$ (a post-hoc convention excluding surpassed targets such as \texttt{OLMo-3-7B} \texttt{MATH-500} and \texttt{Qwen3-4B}/\texttt{Gemma-4-12B-it} \texttt{hh-rlhf}, which remain in $\Delta$AUROC and win counts). $^{\dagger}$At ceiling ($<0.01$). $^*$Low-power ($n=54$/half).}
\label{tab:four_family}
\scriptsize
\setlength{\tabcolsep}{2.1pt}
\begin{tabular}{@{}lcccc|cccccc@{}}
\toprule
 & \multicolumn{4}{c|}{\textbf{Part A: New-18 Out-of-Family Replication}} & \multicolumn{6}{c}{\textbf{Part B: Multi-Source Averaging (6 Targets) \& Combined Recipe}} \\
\cmidrule(lr){2-5}\cmidrule(lr){6-11}
\textbf{Task} & \textbf{\texttt{ms} (Ret.)} & \textbf{\texttt{l2\_lr}($C^*$)} & \textbf{Eig.\ $\rho$} & \textbf{$m(.3)\to m(1)$} & \textbf{Sing.-Med} & \textbf{Best (Or.)} & \textbf{Multi-Src} & \textbf{$\Delta$AUROC (Gap, Win)} & \textbf{Nat.\ \texttt{ms}} & \textbf{Comb.\ 20k$^{\ddagger}$} \\
\midrule
hh-rlhf$^{\dagger}$ & 0.9784 (99.0\%) & 0.9762 & +0.619 & 0.58$\to$0.46 & 0.9788 & 0.9819 & 0.9823 & +0.004 (61.0\%, 6/6) & 0.9824 & 0.9819 \\
XSTest & 0.9650 (96.5\%) & 0.8906 & +0.484 & 0.27$\to$0.30 & 0.9674 & 0.9760 & 0.9761 & +0.006 (45.7\%, 5/6) & 0.9801 & 0.9753 \\
TruthfulQA & 0.8054 (86.6\%) & 0.8020 & +0.447 & 0.17$\to$0.35 & 0.8082 & 0.8307 & 0.8213 & +0.018 (43.5\%, 6/6) & 0.8652 & 0.8375 \\
MATH-500$^*$ & 0.7649 (94.4\%) & 0.6987 & +0.456 & 0.19$\to$0.23 & 0.7710 & 0.7915 & 0.7810 & +0.012 (38.3\%, 6/6) & 0.7802 & 0.7826 \\
\bottomrule
\end{tabular}
\end{table*}

\begin{figure*}[t]
    \centering
    \includegraphics[width=0.74\textwidth]{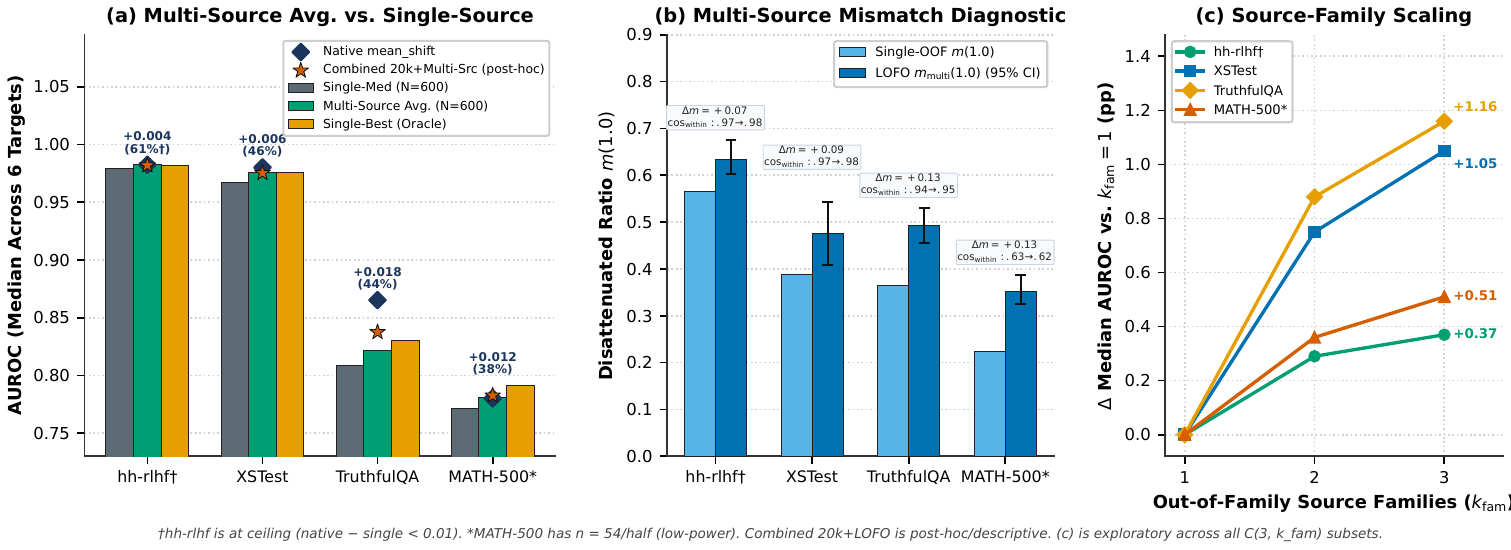}
    \caption{Leave-one-family-out multi-source averaging closes 38\%--46\% of non-ceiling single-source transfer gaps without target labels (6 targets, 4 families, depth $0.6\to 0.6$, 3 seeds). (a) $N=600$ averaging adds $+0.006$--$+0.018$ AUROC ($+0.004$ / 61\% on near-ceiling \texttt{hh-rlhf}); stars mark the post-hoc Combined Best Recipe ($N=20{,}000$ + multi-source). (b) Disattenuated $m_{\mathrm{multi}}(1.0)$ rises $+0.07$ to $+0.13$ over single-source OOF $m(1.0)$ with minimal $\cos_{\mathrm{within}}$ change. (c) Exploratory source-family scaling over $\binom{3}{k_{\mathrm{fam}}}$ subsets.}
    \label{fig:four_family}
\end{figure*}

\subsection{Out-of-Family Replication and Task-Dependent Causes}
\label{sec:oof_rep}

Adding Llama~3.1 and OLMo~3 replicates our core findings across 18 brand-new out-of-family pairs (Table~\ref{tab:four_family}, left; depth 0.5 in Table~\ref{tab:app_v5_05}; parity audit in Appendix~\ref{app:hyperparams}). Strict step-by-step monotonicity held on 2/4 tasks (two within-noise dips occurred on \texttt{MATH-500} before a $+0.099$ jump at $\gamma=1.0$), yet transfer AUROC and matched retention still peaked at mean-shift on all four core tasks (87\%--99\%). Eigenvector alignment met all three pre-registered gates on 18/18 out-of-family pairs (120/120 positive entries across all 30 pairs), and on \texttt{XSTest} mean-shift reached 0.97 AUROC versus 0.89 for source-CV logistic probes and 0.76 for fixed-depth Relative Representations.

Comparing split-half mismatch across slices shows that while mean-shift transferred best on all four tasks, the dominant reason varies by task: \texttt{TruthfulQA} replicated shrinkage-driven mismatch reduction ($0.17 \to 0.35$) and source-specific overfitting, whereas \texttt{hh-rlhf} reversed direction ($0.58 \to 0.46$), source-size Check~\#5 was not met, and \texttt{XSTest} reflected source-CV domain shift on \texttt{JBB}.

\subsection{Leave-One-Family-Out Multi-Source Averaging}
\label{sec:lofo}

Averaging rotated mean-shift probes across out-of-family sources closes 38\%--46\% of non-ceiling single-source transfer gaps without target labels, meeting all four pre-registered criteria (Table~\ref{tab:four_family}, right; Figure~\ref{fig:four_family}a; per-target breakdown in Table~\ref{tab:app_v5_pertarget}). Unweighted and family-balanced averaging agree within 0.002 AUROC on every task. Excluding targets where single-source transfer already surpasses native performance (a post-hoc convention for the gap-closed ratio median), multi-source averaging gains $+0.006$ on \texttt{XSTest} (46\% gap closed, 5/6 targets), $+0.018$ median on \texttt{TruthfulQA} (44\%, 6/6), and $+0.012$ on \texttt{MATH-500} (38\%, 6/6), matching the oracle single source on two tasks and staying within 0.011 on the other two without target labels.

Measuring disattenuated alignment in target space shows $m_{\mathrm{multi}}(1.0)$ rising $+0.07$--$+0.13$ over single-source pairs across all four tasks with minimal reliability change (Figure~\ref{fig:four_family}b), consistent with cross-family mismatch cancellation. Three companion checks refine deployment design: pooled cross-source logistic regression reaches 0.85 on \texttt{TruthfulQA} yet trails mean-shift averaging on \texttt{XSTest} and \texttt{MATH-500} (Table~\ref{tab:app_v5_pertarget}); adding a second source family brings larger gains ($+0.003$--$+0.011$) than adding a third ($+0.0006$--$+0.0036$; Figure~\ref{fig:four_family}c); and combining 20,000 anchors with multi-source averaging lifts \texttt{TruthfulQA} to 0.838 (92\% retention, 51\% gap closed; Table~\ref{tab:four_family}). The Combined Best Recipe is a post-hoc, descriptive summary, not a pre-registered test.

\section{Limitations, Practical Guidance, and Conclusion}
\label{sec:limitations_guidance}

\paragraph{Limitations.}
Seven boundaries qualify our findings: (1) models span 4B--12B parameters, and four-family coverage spans the four core tasks, map-side, multi-layer Relative Representations, anchor-bank, and clean \texttt{deepset} studies, while secondary tasks span two families (Turkish prompt injection was translated via Gemma-3 and evaluated on two Gemma targets); (2) our 100 trigger-word hard negatives are author-audited synthetic prompts rather than production logs; (3) short-trained probes degrade on 4K indirect injections (cause unestablished) and adaptive evasion remains open; (4) directed pairs share six models (limiting small-cluster bootstrap coverage), though LOMO, LOFO, two-way (source $\times$ target), and 1,000-resample paired test-item / pair $\times$ item bootstraps confirm positive intervals across all major four-task, multi-layer, map-side, and anchor-diversity margins, while narrow single-task margins cross zero on ceiling \texttt{hh-rlhf}, \texttt{TruthfulQA} \texttt{ms}$-$\texttt{lr}, \texttt{deepset} in-domain vs.\ mixed, and finite-sample \texttt{MATH-500} ($n=108$) in-domain gains (Table~\ref{tab:app_v7_a5}); (5) Behavioral Sycophancy covers 32 claims across four models (Appendix~\ref{app:hyperparams}); (6) \texttt{MATH-500} ($n=108$) and \texttt{deepset} ($n=116$) have modest test splits (paired item CIs in Table~\ref{tab:app_v7_a5}); and (7) mixed anchors were verified up to 5,000 prompts and in-domain anchors up to 600 prompts (including template-matched \texttt{TruthfulQA}), while 20,000-anchor scaling used \texttt{SQuAD} prose and specialized domains (biomedical, legal) remain untested.

\paragraph{Practical Guidance.}
Five rules summarize portable linear monitoring:
\begin{enumerate}
    \item \textit{Default to mean-shift probes ($\gamma=1.0$), noting lexical-shift exceptions:} Mean-shift probes ($\gamma = 1.0$) achieved the highest transfer across the four core tasks ($+0.054$ over source-CV logistic probes, two-way CI $>0$). On \texttt{xTRam1} ($n=392$) $\to$ \texttt{deepset}, source-centered logistic probes rise from 0.835 to 0.944 with mixed anchors and 0.956 with in-domain anchors (vs.\ 0.781 to 0.887/0.914 mean-shift and 0.79 MiniLM; Table~\ref{tab:v8_mixed}).
    \item \textit{Avoid source-model cross-validation for transfer regularization,} which under-regularizes and degrades cross-model transfer.
    \item \textit{Calibrate thresholds on in-domain benign traffic and compare unsafe recall at fixed false-alarm rate:} While $\approx 5.0\%$ held-out false alarms hold by exchangeability, transferred mean-shift reaches 81.1\% unsafe recall vs.\ 50.7\% source-CV logistic, 6.2\% Bakker logistic, and 40.6\% Bakker mean-shift (Tables~\ref{tab:app_v7_a2} and~\ref{tab:v8_bakker}), whereas OOD calibration inflates false alarms to 31\%.
    \item \textit{Use transfer for zero-label cold-start or unshareable-corpus upgrades; retrain natively when source texts exist:} Single-source transfer (600 anchors) matches $k^* \le 4$--$57$ target labels, rising to $6$ to $>64$ with multi-source averaging (Figure~\ref{fig:sample_eff}a, Table~\ref{tab:app_v7_a4}); target labels remain needed for behavioral tasks.
    \item \textit{Diversify small anchor banks, use unlabeled in-domain anchors when available, and average multiple sources:} Balanced 600-prompt four-domain anchors lift \texttt{deepset} OOD transfer by $+0.10$ without per-task rotations, 600 unlabeled in-domain prompts add $+0.02$ across core tasks ($+0.045$ on \texttt{TruthfulQA}), and multi-source averaging closes 38\%--46\% of non-ceiling gaps.
\end{enumerate}

\paragraph{Conclusion.}
Across six dense and hybrid models from four families, read-only linear probe transfer retains high above-chance AUROC on input-property monitoring yet degrades on behavioral sycophancy (where shared counter-bias mean-shift drops reflect label disagreement rather than direction-transfer failure), long-context injections, and iterative evasion. Folded depth-matched Orthogonal Procrustes beats fixed-depth and full multi-layer global Relative Representations at a fraction of serving memory and latency. Across our four core tasks, mean-shift probes ($\gamma = 1.0$) achieved the highest transfer ($+0.054$ over source-CV logistic probes, two-way CI $>0$) for task-dependent reasons, while Orthogonal Procrustes led whitened, Ridge, Kernel Ridge, and neural maps despite lower validation $R^2$ and cosine because anisotropic shrinkage distorts target directions and lowers mean test signal-to-noise ratio. Strong transfer does not require tight directional agreement: on \texttt{MATH-500}, rotated and native probes agree at cosine roughly 0.13 yet retain roughly 95\% of native above-chance AUROC. Balanced four-domain anchors boost out-of-distribution robustness without per-task rotations, unlabeled in-domain anchors add further gains when target traffic is available, multi-source averaging closes transfer gaps without target labels, and held-out in-domain calibration yields strong unsafe recall at false-alarm rates near nominal targets.

\paragraph{AI Usage Disclosure.}
{\footnotesize AI coding and writing assistants (including RuneLab's internal research assistant \href{https://runelab.ai/en/snotra}{Snotra-Science-1}) were used under human author direction to help implement experiment and plotting scripts and edit manuscript prose. The human authors designed the research program, directed all analyses, audited all experimental outputs and statistical tables, and take full responsibility for the paper.\par}

\clearpage
{\footnotesize
\let\oldthebibliography\thebibliography
\renewcommand\thebibliography[1]{\oldthebibliography{#1}\setlength{\itemsep}{0.5pt}\setlength{\parsep}{0pt}\setlength{\parskip}{0pt}}
\bibliographystyle{plain}
\bibliography{references}
}

\clearpage
\onecolumn
\raggedbottom
\appendix
\setcounter{totalnumber}{6}
\setcounter{topnumber}{6}
\setcounter{bottomnumber}{6}
\renewcommand{\topfraction}{0.95}
\renewcommand{\bottomfraction}{0.95}
\renewcommand{\textfraction}{0.03}
\setlength{\floatsep}{2.8pt plus 0.6pt minus 0.6pt}
\setlength{\textfloatsep}{2.8pt plus 0.6pt minus 0.6pt}
\setlength{\intextsep}{2.5pt plus 0.6pt minus 0.6pt}
\setlength{\abovecaptionskip}{1.8pt}
\setlength{\belowcaptionskip}{1.8pt}
\renewcommand{\arraystretch}{0.95}
\titlespacing*{\section}{0pt}{3.8pt plus 0.6pt minus 0.6pt}{1.5pt plus 0.5pt}

\section{Secondary Depth \texorpdfstring{($0.5 \to 0.5$)}{(0.5 -> 0.5)} Sweeps, Four-Family, and Map-Family Checks}
\label{app:depth05}

Tables~\ref{tab:app_cov_05}, \ref{tab:app_mech_05}, \ref{tab:app_v5_05}, and~\ref{tab:app_v6_05} report secondary depth $0.5 \to 0.5$ results corresponding to Sections~\ref{sec:probe_side}--\ref{sec:four_family}. Near-monotone transfer gains with increasing $\gamma$ (within 0.005 at $\gamma=1.0$ on \texttt{XSTest} and \texttt{TruthfulQA}), source-CV under-regularization, positive eigenvector alignment across all pairs (48/48 Original~12; 72/72 New~18), Whitened-Rotation degradation on all four core tasks (Table~\ref{tab:app_v6_05}), and multi-source averaging gains all replicate at depth 0.5, with lower overall transfer AUROC than depth 0.6 on \texttt{TruthfulQA} and \texttt{XSTest}. On \texttt{XSTest} at depth 0.5 (Original~12), transferred \texttt{mean\_shift} safe-prompt false-alarm rates at 50\%, 70\%, 80\%, and 90\% label-dependent matched unsafe recall are 0.8\%, 4.5\%, 10.6\%, and 23.5\%, versus 4.5\%, 10.8\%, 17.1\%, and 25.0\% for transferred \texttt{l2\_lr}($C^*$).

\begin{table}[htbp]
\centering
\caption{Secondary depth $0.5\to 0.5$ shrinkage-LDA sweep and source 5-fold CV \texttt{l2\_lr}($C^*$) (Original-12 median, 3 seeds; default centering, shift-invariant / $\mu_{\mathrm{anc}}$).}
\label{tab:app_cov_05}
\footnotesize
\setlength{\tabcolsep}{3.4pt}
\begin{tabular}{@{}lccccccc@{}}
\toprule
\textbf{Task} & \textbf{$\gamma=0.001$} & \textbf{$\gamma=0.1$} & \textbf{$\gamma=0.5$} & \textbf{$\gamma=0.9$} & \textbf{$\gamma=1.0$ (\texttt{ms})} & \textbf{Matched Ret.} & \textbf{Source CV \texttt{l2\_lr}($C^*$)} \\
 & \textbf{Trans / Nat} & \textbf{Trans / Nat} & \textbf{Trans / Nat} & \textbf{Trans / Nat} & \textbf{Trans / Nat} & \textbf{$\gamma:0.001\to 1.0$} & \textbf{Trans ($\cos(w,\Delta\mu)$)} \\
\midrule
Anthropic/hh-rlhf & 0.858 / 0.956 & 0.875 / 0.961 & 0.920 / 0.973 & 0.961 / 0.978 & 0.972 / 0.973 & 77.8\%$\to$99.3\% & 0.9684 (0.52) \\
Paul/XSTest (JBB) & 0.710 / 0.915 & 0.739 / 0.919 & 0.828 / 0.935 & 0.934 / 0.959 & 0.931 / 0.967 & 51.1\%$\to$96.6\% & 0.9028 (0.34) \\
TruthfulQA (QID) & 0.500 / 0.792 & 0.512 / 0.848 & 0.599 / 0.884 & 0.672 / 0.886 & 0.667 / 0.808 & -0.7\%$\to$59.1\% & 0.6252 (0.50) \\
MATH-500 ($\pm 15\%$) & 0.489 / 0.715 & 0.491 / 0.718 & 0.519 / 0.730 & 0.603 / 0.760 & 0.762 / 0.769 & -6.1\%$\to$98.4\% & 0.6465 (0.48) \\
\bottomrule
\end{tabular}
\end{table}

\begin{table}[htbp]
\centering
\caption{Secondary depth $0.5\to 0.5$ eigenvector alignment and split-half decomposition (Original~12 pairs; test activations centered by $\mu_{\mathrm{anc}}$, shrinkage-LDA shift-invariant). Variance preservation ratio is $\mathrm{VPR}_k = (\operatorname{Var}(X_{B,\mathrm{test}} R v_k) / \operatorname{Var}(X_{A,\mathrm{test}} v_k)) / (\|B_0\|_F^2 / \|A_0\|_F^2)$.}
\label{tab:app_mech_05}
\footnotesize
\setlength{\tabcolsep}{3.6pt}
\begin{tabular}{@{}lcccccc@{}}
\toprule
\textbf{Task} & \textbf{Eig.\ $\rho$ ($>0$)} & \textbf{$\Delta$(Q1$-$Q4)} & \textbf{VPR (Q1 / Q4)} & \textbf{$\cos_{\mathrm{within}}$ ($0.3\to 1.0$)} & \textbf{Span Loss (1.0)} & \textbf{$m(0.3)\to m(1.0)$} \\
\midrule
Anthropic/hh-rlhf & +0.4695 (12/12) & +0.2217 & 0.66 / 2.74 & 0.101$\to$0.974 & 6.0\% & 0.175$\to$0.512 \\
Paul/XSTest (JBB) & +0.3923 (12/12) & +0.2576 & 0.42 / 1.33 & 0.182$\to$0.963 & 11.3\% & 0.199$\to$0.408 \\
TruthfulQA (QID) & +0.3748 (12/12) & +0.1545 & 0.32 / 2.05 & 0.184$\to$0.909 & 23.5\% & 0.106$\to$0.154 \\
MATH-500 ($\pm 15\%$) & +0.3319 (12/12) & +0.1846 & 0.09 / 0.65 & 0.075$\to$0.757 & 23.8\% & 0.240$\to$0.171 \\
\bottomrule
\end{tabular}
\end{table}

\begin{table}[htbp]
\centering
\caption{Secondary depth $0.5\to 0.5$ four-family results: New-18 single-source replication and leave-one-family-out multi-source averaging (6-target median; default centering, shift-invariant / $\mu_{\mathrm{anc}}$).}
\label{tab:app_v5_05}
\footnotesize
\setlength{\tabcolsep}{3.4pt}
\begin{tabular}{@{}lccccccc@{}}
\toprule
\textbf{Task} & \textbf{New-18 \texttt{ms} (Ret.)} & \textbf{New-18 \texttt{l2\_lr}($C^*$)} & \textbf{Eig.\ $\rho$ ($>0$)} & \textbf{LOFO Sing.-Med} & \textbf{Multi-Src} & \textbf{Nat.\ \texttt{ms}} & \textbf{Wins} \\
\midrule
hh-rlhf$^{\dagger}$ & 0.9741 (98.5\%) & 0.9696 & +0.519 (18/18) & 0.9716 & 0.9780 & 0.9780 & 6/6 \\
Paul/XSTest & 0.9570 (96.7\%) & 0.9013 & +0.429 (18/18) & 0.9375 & 0.9591 & 0.9734 & 6/6 \\
TruthfulQA & 0.7617 (75.8\%) & 0.7254 & +0.392 (18/18) & 0.7441 & 0.7986 & 0.8559 & 6/6 \\
MATH-500$^*$ & 0.7393 (84.7\%) & 0.6608 & +0.339 (18/18) & 0.7361 & 0.7743 & 0.7725 & 6/6 \\
\bottomrule
\end{tabular}
\end{table}

\begin{table}[htbp]
\centering
\caption{Secondary depth $0.5\to 0.5$ map-family check at $N=600$ (All-30 pairs, 3 seeds, shift-invariant \texttt{mean\_shift}; maps fit on $\mu_{\mathrm{anc}}$-centered anchors). Whitened Rotation (\texttt{M1}) again achieves higher CV anchor cosine than Procrustes yet trails on all 4 core tasks.}
\label{tab:app_v6_05}
\footnotesize
\setlength{\tabcolsep}{4.2pt}
\begin{tabular}{@{}lcccccc@{}}
\toprule
\textbf{Map Family ($N=600$, Depth $0.5\to 0.5$)} & \textbf{CV Cos} & \textbf{hh-rlhf} & \textbf{XSTest} & \textbf{TruthfulQA} & \textbf{MATH-500} & \textbf{4-Task Mean} \\
\midrule
M0 Orthogonal Procrustes (Ref) & 0.5533 & 0.9730 & 0.9431 & 0.7085 & 0.7473 & 0.8430 \\
M1 Whitened Rotation (CV $\varepsilon$) & 0.6284 & 0.9372 & 0.9032 & 0.6324 & 0.6572 & 0.7825 \\
M2 Ridge Regression (CV $\lambda$) & 0.6963 & 0.9612 & 0.9301 & 0.6823 & 0.7259 & 0.8249 \\
M5 Kernel Ridge RBF (CV) & 0.6972 & 0.9576 & 0.8719 & 0.6956 & 0.7319 & 0.8143 \\
\bottomrule
\end{tabular}
\end{table}

\section{Per-Pair, Per-Target, Supplementary Figures, Calibration, and Map-Side Breakdowns \texorpdfstring{($0.6 \to 0.6$)}{(0.6 -> 0.6)}}
\label{app:pairs}

\begin{figure}[htbp]
    \centering
    \includegraphics[width=0.78\textwidth]{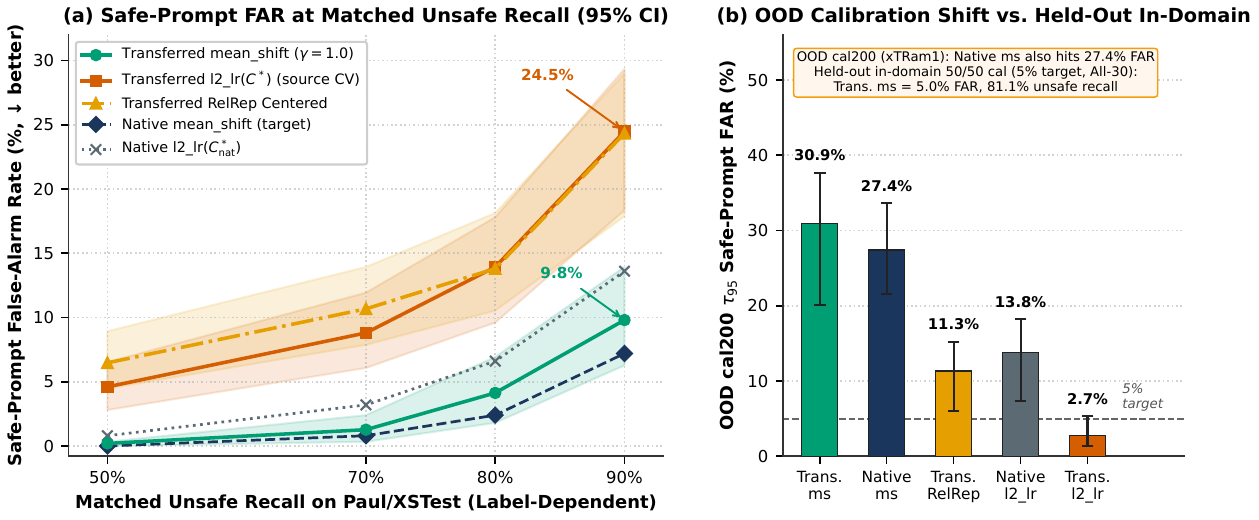}
    \caption{Safe-prompt false-alarm operating points on \texttt{XSTest} (trained on \texttt{JBB}, depth $0.6\to 0.6$, Original~12 pairs; probe alarms, not model refusals). (a) At label-dependent matched unsafe recall (50\%--90\%), transferred \texttt{mean\_shift} attains 0.2\%--9.8\% safe-prompt false-alarm rate, below \texttt{l2\_lr}($C^*$) (4.6\%--24.5\%) and fixed-depth centered Relative Representations (6.5\%--24.3\%). (b) Under out-of-domain \texttt{cal200} ($\tau_{95}$ on \texttt{xTRam1} benign prompts), both transferred (30.9\%) and native (27.4\%) \texttt{mean\_shift} trigger high false-alarm rates, whereas held-out in-domain 50/50 calibration yields 81.1\% unsafe recall for transferred \texttt{mean\_shift} at 5.0\% calibrated safe-prompt FAR (vs.\ 50.7\% \texttt{l2\_lr}($C^*$) and 17.7\% fixed-depth centered RelRep; Table~\ref{tab:app_v7_a2}).}
    \label{fig:xstest_cal}
\end{figure}

\begin{figure}[htbp]
    \centering
    \includegraphics[width=0.80\textwidth]{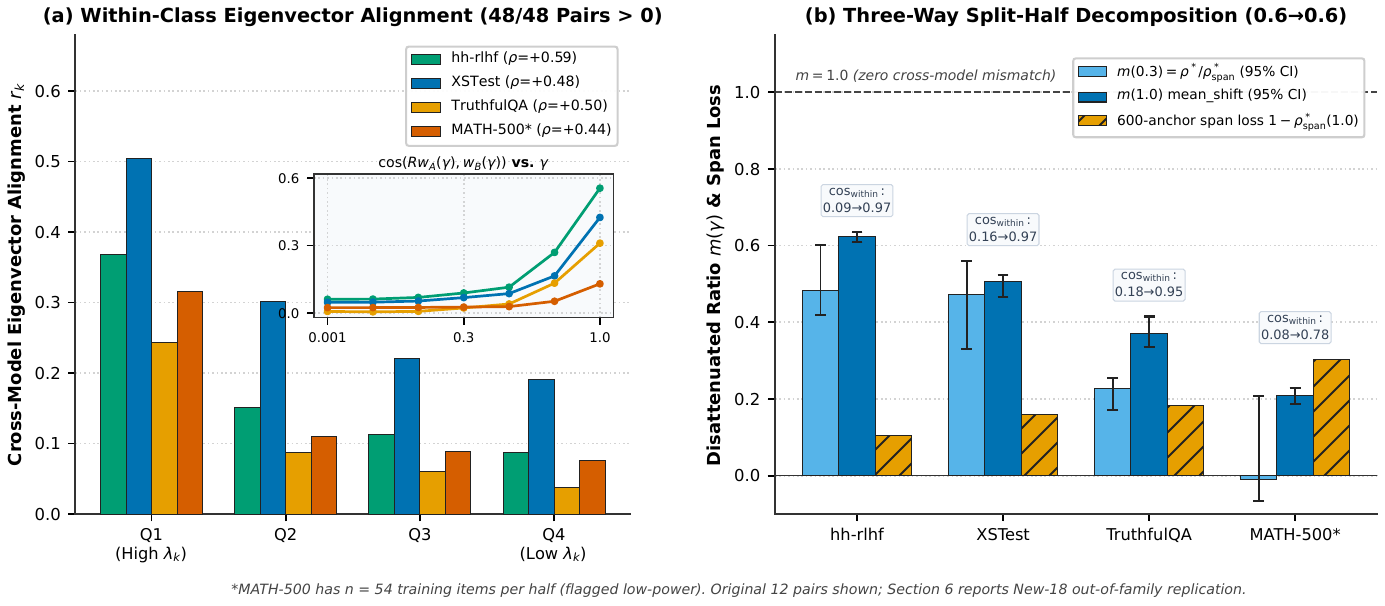}
    \caption{Probe-side mechanism at depth $0.6\to 0.6$ (Original~12 pairs). (a) Cross-model alignment $r_k$ of within-class covariance eigenvectors by eigenvalue quartile (Q1 high to Q4 low); 48/48 pair-task entries have positive $\rho(\log \lambda_k, r_k)$, and probe cosine rises with $\gamma$ (inset). (b) Split-half disattenuated decomposition separating within-model reliability ($\cos_{\mathrm{within}}$), 600-anchor span loss ($1-\rho^*_{\mathrm{span}}$), and cross-model mismatch ratio $m(\gamma)=\rho^*/\rho^*_{\mathrm{span}}$ (95\% CIs).}
    \label{fig:mechanism}
\end{figure}

Figures~\ref{fig:xstest_cal} and~\ref{fig:mechanism} present the \texttt{XSTest} label-dependent matched-recall curves, \texttt{cal200} calibration-domain shift audit, within-class eigenvector alignment quartiles, and three-way split-half decomposition supporting Section~\ref{sec:probe_side}. Table~\ref{tab:app_v7_a2} reports the canonical held-out in-domain 50/50 calibration audit on \texttt{XSTest}. Table~\ref{tab:app_pairs_eig} lists all 12 directed Original-12 cross-model pairs for within-class eigenvector alignment $\rho(\log \lambda_k, r_k)$ and quartile gap $\Delta(\text{Q1}-\text{Q4})$ at primary depth $0.6 \to 0.6$. Table~\ref{tab:app_v5_pertarget} lists the per-target 6-model multi-source breakdown alongside the post-hoc Combined Best Recipe ($N=20{,}000$ + multi-source) and 6-target median \texttt{LOFO-Pooled-L2LR}. Tables~\ref{tab:app_v6_mapfamily}, \ref{tab:app_v6_scaling_4tasks}, and~\ref{tab:app_v7_a1} report the canonical map-family comparison at $N=600$, the $N \in \{600, 2\text{k}, 5\text{k}, 20\text{k}\}$ scaling and CPU serving benchmarks, and the Scaled Procrustes decomposition strictly on the four core tasks. Tables~\ref{tab:v8_bakker}, \ref{tab:v8_mixed}, and~\ref{tab:v8_map_mech} report the canonical All-30 head-to-head against full multi-layer Bakker and Louizos~\cite{bakker2026relative} global angular-distance Relative Representations (with \texttt{OMP\_NUM\_THREADS=4} serving benchmarks), the balanced four-domain, unlabeled in-domain, and 1D score-distillation comparison, and the map-side dissociation mechanism (probe-direction cosine, held-out test SNR, and Ridge singular-value shrinkage).

\begin{table}[htbp]
\centering
\caption{Held-out in-domain unsafe recall at calibrated safe-prompt FAR on \texttt{XSTest} across 100 random 50/50 splits of the 250 safe prompts (125 calibration / 125 held-out test; recall on 200 unsafe prompts, depth $0.6\to 0.6$, 3 anchor seeds, unified canonical All-30 pair-medians; default centering, shift-invariant / $\mu_{\mathrm{anc}}$). Thresholds use the empirical $(1-\alpha)$-quantile on the calibration half (final column: exact split-conformal order statistic). Under i.i.d.\ 50/50 splits, achieving $\approx 5.0\%$ held-out FAR at a 5\% target holds by exchangeability for any score function (meeting our pre-registered $\le 8\%$ check); the discriminative separation lies in unsafe recall at that fixed $\approx 5\%$ FAR: 81.12\% for transferred \texttt{mean\_shift} vs.\ 50.71\% for \texttt{l2\_lr}($C^*$), 17.67\% for anchor-centered fixed-depth RelRep (28.0\% Orig-12, 17.5\% New-18), 6.2\% for Bakker raw \texttt{l2\_lr}($C^*$) (40.6\% Bakker \texttt{mean\_shift}; exact matched run in Table~\ref{tab:v8_bakker}), and 86.00\% for native \texttt{mean\_shift}.}
\label{tab:app_v7_a2}
\scriptsize
\setlength{\tabcolsep}{3.0pt}
\begin{tabular}{@{}lccccc@{}}
\toprule
\textbf{Probe / Slice} & \textbf{Target} & \textbf{Held-Out Safe FAR [IQR]} & \textbf{FAR p95} & \textbf{Unsafe Recall [IQR]} & \textbf{Conformal FAR / Rec.} \\
\midrule
Transferred \texttt{mean\_shift} (All-30) & 1\% & 1.33\% [1.1, 1.4] & 4.5\% & 69.92\% [62.0, 73.7] & 0.00\% / 47.7\% \\
Transferred \texttt{mean\_shift} (All-30) & 5\% & 5.00\% [4.8, 5.3] & 10.2\% & 81.12\% [78.4, 85.2] & 4.00\% / 79.3\% \\
Transferred \texttt{mean\_shift} (All-30) & 10\% & 9.87\% [9.6, 10.1] & 17.1\% & 89.50\% [86.4, 91.6] & 8.80\% / 88.1\% \\
\midrule
Transferred \texttt{mean\_shift} (Orig-12) & 5\% & 4.87\% [4.8, 5.3] & 9.9\% & 82.42\% [80.0, 84.8] & 4.00\% / 80.5\% \\
Transferred \texttt{mean\_shift} (New-18) & 5\% & 5.07\% [4.8, 5.3] & 10.4\% & 79.50\% [75.1, 87.0] & 4.00\% / 78.1\% \\
Transferred \texttt{l2\_lr}($C^*$) (All-30) & 5\% & 5.27\% [5.1, 5.3] & 10.4\% & 50.71\% [28.7, 62.9] & 4.27\% / 48.7\% \\
Transferred RelRep Cen.\ (All-30, depth 0.6) & 5\% & 5.33\% [4.9, 5.3] & 10.9\% & 17.67\% [1.8, 48.0] & 4.20\% / 15.5\% \\
Native \texttt{mean\_shift} (6 models) & 5\% & 4.80\% [4.8, 4.8] & 10.0\% & 86.00\% [84.6, 88.9] & 4.00\% / 85.8\% \\
Native \texttt{l2\_lr}($C^*$) (6 models) & 5\% & 5.00\% [4.8, 5.5] & 10.4\% & 80.50\% [77.4, 83.6] & 4.00\% / 78.8\% \\
\bottomrule
\end{tabular}
\end{table}

\begin{table}[htbp]
\centering
\caption{Complete Original-12 per-pair within-class eigenvector alignment ($\rho(\log\lambda_k, r_k)$ and $\Delta(\text{Q1}-\text{Q4})$) at depth $0.6\to 0.6$ (held-out test activations centered by $\mu_{\mathrm{anc}}$). All 48/48 Original-12 pairs (and 72/72 New-18 pairs; 120/120 total) exhibit positive $\rho$ and $\Delta(\text{Q1}-\text{Q4})\ge +0.15$.}
\label{tab:app_pairs_eig}
\footnotesize
\setlength{\tabcolsep}{3.6pt}
\begin{tabular}{@{}lcccccccc@{}}
\toprule
\textbf{Source $\to$ Target} & \multicolumn{2}{c}{\textbf{hh-rlhf}} & \multicolumn{2}{c}{\textbf{XSTest}} & \multicolumn{2}{c}{\textbf{TruthfulQA}} & \multicolumn{2}{c}{\textbf{MATH-500}} \\
\cmidrule(lr){2-3}\cmidrule(lr){4-5}\cmidrule(lr){6-7}\cmidrule(lr){8-9}
 & \textbf{$\rho$} & \textbf{$\Delta$(Q1$-$Q4)} & \textbf{$\rho$} & \textbf{$\Delta$(Q1$-$Q4)} & \textbf{$\rho$} & \textbf{$\Delta$(Q1$-$Q4)} & \textbf{$\rho$} & \textbf{$\Delta$(Q1$-$Q4)} \\
\midrule
Qwen3-4B $\to$ Qwen3.5-9B & +0.612 & +0.284 & +0.404 & +0.273 & +0.501 & +0.203 & +0.567 & +0.328 \\
Qwen3-4B $\to$ Gemma-4-E4B-it & +0.526 & +0.271 & +0.437 & +0.256 & +0.439 & +0.184 & +0.520 & +0.289 \\
Qwen3-4B $\to$ Gemma-4-12B-it & +0.506 & +0.250 & +0.349 & +0.215 & +0.373 & +0.157 & +0.329 & +0.155 \\
Qwen3.5-9B $\to$ Qwen3-4B & +0.606 & +0.266 & +0.483 & +0.291 & +0.478 & +0.201 & +0.473 & +0.263 \\
Qwen3.5-9B $\to$ Gemma-4-E4B-it & +0.660 & +0.329 & +0.544 & +0.281 & +0.516 & +0.232 & +0.552 & +0.287 \\
Qwen3.5-9B $\to$ Gemma-4-12B-it & +0.577 & +0.262 & +0.509 & +0.281 & +0.496 & +0.211 & +0.398 & +0.229 \\
Gemma-4-E4B-it $\to$ Qwen3-4B & +0.548 & +0.262 & +0.398 & +0.251 & +0.493 & +0.198 & +0.401 & +0.221 \\
Gemma-4-E4B-it $\to$ Qwen3.5-9B & +0.646 & +0.297 & +0.484 & +0.309 & +0.561 & +0.237 & +0.521 & +0.265 \\
Gemma-4-E4B-it $\to$ Gemma-4-12B-it & +0.575 & +0.296 & +0.545 & +0.359 & +0.514 & +0.228 & +0.333 & +0.174 \\
Gemma-4-12B-it $\to$ Qwen3-4B & +0.518 & +0.260 & +0.411 & +0.247 & +0.434 & +0.187 & +0.388 & +0.204 \\
Gemma-4-12B-it $\to$ Qwen3.5-9B & +0.602 & +0.290 & +0.480 & +0.290 & +0.527 & +0.232 & +0.411 & +0.219 \\
Gemma-4-12B-it $\to$ Gemma-4-E4B-it & +0.624 & +0.317 & +0.516 & +0.332 & +0.526 & +0.241 & +0.611 & +0.319 \\
\bottomrule
\end{tabular}
\end{table}

\begin{table}[htbp]
\centering
\caption{Per-target 6-model LOFO multi-source transfer at depth $0.6\to 0.6$ (shift-invariant \texttt{mean\_shift}; \texttt{Pool-LR} uses $\mu_{\mathrm{anc}}$-centered rotated states): Single-Med ($N=600$) / Multi-Source Avg.\ (MS600) ($N=600$) / Post-Hoc Combined ($N=20{,}000$ + LOFO) / Native \texttt{mean\_shift}. Bottom row reports 6-target median \texttt{LOFO-Pooled-L2LR}.}
\label{tab:app_v5_pertarget}
\scriptsize
\setlength{\tabcolsep}{2.4pt}
\begin{tabular}{@{}lcccc@{}}
\toprule
\textbf{Target} & \textbf{hh-rlhf (S600/MS600/C20k/Nat)} & \textbf{XSTest (S600/MS600/C20k/Nat)} & \textbf{TruthfulQA (S600/MS600/C20k/Nat)} & \textbf{MATH-500 (S600/MS600/C20k/Nat)} \\
\midrule
Qwen3-4B & 0.980/0.983/0.982/0.977 & 0.946/0.945/0.963/0.967 & 0.788/0.815/0.828/0.870 & 0.779/0.782/0.787/0.786 \\
Qwen3.5-9B & 0.979/0.985/0.985/0.984 & 0.961/0.964/0.974/0.978 & 0.852/0.869/0.882/0.875 & 0.770/0.780/0.778/0.774 \\
Gemma-4-E4B-it & 0.976/0.978/0.981/0.981 & 0.965/0.973/0.973/0.974 & 0.809/0.822/0.841/0.840 & 0.777/0.839/0.799/0.813 \\
Gemma-4-12B-it & 0.981/0.982/0.982/0.980 & 0.971/0.979/0.977/0.982 & 0.841/0.871/0.876/0.907 & 0.762/0.772/0.777/0.788 \\
Llama-3.1-8B & 0.973/0.978/0.976/0.984 & 0.984/0.986/0.993/0.993 & 0.807/0.820/0.834/0.860 & 0.732/0.747/0.754/0.771 \\
OLMo-3-7B & 0.978/0.982/0.984/0.984 & 0.970/0.985/0.989/0.983 & 0.781/0.801/0.791/0.821 & 0.772/0.814/0.806/0.771 \\
\midrule
Med.\ Pool-LR & \texttt{Pool-LR}: 0.9780 & \texttt{Pool-LR}: 0.9216 & \texttt{Pool-LR}: 0.8468 & \texttt{Pool-LR}: 0.7410 \\
\bottomrule
\end{tabular}
\end{table}

\begin{table}[htbp]
\centering
\caption{Canonical map-family comparison at $N=600$ \texttt{SQuAD} anchors (All-30 pair-medians, 3 seeds, depth $0.6\to 0.6$, 4 core tasks; default centering, $\mu_{\mathrm{anc}}$ / shift-invariant). Difference $\Delta$ vs.\ \texttt{M0} shows paired-median deltas with pair-cluster 95\% bootstrap CIs (two-way source$\times$target 4-task CIs exclude zero for all five alternatives \texttt{M1}--\texttt{M5}; Table~\ref{tab:app_v7_a5}). For \texttt{M1}, 5-fold CV selected $\varepsilon_{\mathrm{rel}}=10^{-2}$ on 56/90 fits and $10^{-3}$ on 34/90 fits, meeting our pre-registered Rule~A on 3/4 tasks (\texttt{MATH-500} upper CI $+0.0002$; 4/4 in archived depth-0.5 replication, Table~\ref{tab:app_v6_05}).}
\label{tab:app_v6_mapfamily}
\scriptsize
\setlength{\tabcolsep}{2.2pt}
\resizebox{\textwidth}{!}{%
\begin{tabular}{@{}lccccccc@{}}
\toprule
\textbf{Map ($N=600$)} & \textbf{CV Cos} & \textbf{hh-rlhf ($\Delta$ [95\% CI])} & \textbf{XSTest ($\Delta$ [95\% CI])} & \textbf{TruthfulQA ($\Delta$ [95\% CI])} & \textbf{MATH-500 ($\Delta$ [95\% CI])} & \textbf{4T \texttt{ms}} & \textbf{4T \texttt{lr}($C^*$)} \\
\midrule
M0 Procrustes (Ref) & 0.5982 & 0.9782 (ref) & 0.9662 (ref) & 0.8057 (ref) & 0.7711 (ref) & 0.8783 & 0.8205 \\
M1 Whitened Rot.\ (CV $\varepsilon$) & 0.6774 & 0.964 (-0.010 [-0.017,-0.008]) & 0.942 (-0.019 [-0.035,-0.007]) & 0.753 (-0.059 [-0.090,-0.048]) & 0.755 (-0.010 [-0.022,+0.000]) & 0.8514 & 0.8113 \\
M2 Ridge (CV $\lambda$) & 0.7112 & 0.967 (-0.012 [-0.017,-0.007]) & 0.934 (-0.025 [-0.047,-0.013]) & 0.738 (-0.060 [-0.087,-0.046]) & 0.757 (-0.014 [-0.027,-0.006]) & 0.8462 & 0.8110 \\
M3 Proc.+ResMLP (CV) & 0.6584 & 0.967 (-0.011 [-0.015,-0.005]) & 0.941 (-0.017 [-0.034,-0.007]) & 0.753 (-0.050 [-0.065,-0.035]) & 0.756 (-0.012 [-0.018,-0.003]) & 0.8531 & 0.8065 \\
M4 MLP Scratch (CV) & 0.6658 & 0.953 (-0.027 [-0.041,-0.018]) & 0.903 (-0.053 [-0.119,-0.022]) & 0.694 (-0.121 [-0.146,-0.103]) & 0.721 (-0.042 [-0.057,-0.029]) & 0.8118 & 0.7775 \\
M5 Kernel Ridge RBF (CV) & 0.7120 & 0.966 (-0.012 [-0.019,-0.007]) & 0.892 (-0.071 [-0.133,-0.041]) & 0.746 (-0.061 [-0.077,-0.042]) & 0.751 (-0.011 [-0.031,+0.002]) & 0.8274 & 0.8200 \\
\bottomrule
\end{tabular}
}
\end{table}

\begin{table}[htbp]
\centering
\caption{Anchor-count scaling (All-30 pairs, depth $0.6\to 0.6$) strictly on the 4 core tasks, including Scaled Procrustes (\texttt{M0s}, single positive scalar $s^*$ fit on fit anchors; leaves transfer AUROC unchanged by construction via scale invariance, verified to 0.0 in \texttt{float64}) using nested 100\% \texttt{SQuAD} anchors ($N\in\{600, 2\text{k}, 5\text{k}, 20\text{k}\}$, 1,000 fixed validation anchors; default centering, $\mu_{\mathrm{anc}}$ / shift-invariant). Serving cost is CPU 4-thread single-prompt p95 latency at $d=4096$ (KRR exact $N=600$; Nystr\"om $m=2{,}000$ takes 20.5\,ms).}
\label{tab:app_v6_scaling_4tasks}
\scriptsize
\setlength{\tabcolsep}{2.8pt}
\begin{tabular}{@{}lccccccccc@{}}
\toprule
\textbf{Map Family} & \textbf{Val $R^2$ (20k)} & \textbf{Val Cos (20k)} & \textbf{$N=600$} & \textbf{$N=2\text{k}$} & \textbf{$N=5\text{k}$} & \textbf{$N=20\text{k}$ (\texttt{ms})} & \textbf{$N=20\text{k}$ (\texttt{lr}$C^*$)} & \textbf{Serving p95} & \textbf{Runtime Params} \\
\midrule
M0 Orthogonal Procrustes & 0.2274 & 0.6472 & 0.8823 & 0.8884 & 0.8900 & 0.8911 & 0.8626 & 0.0027\,ms & 4,096 \\
M0s Scaled Procrustes ($s^*$) & 0.4105 & 0.6473 & 0.8824 & 0.8884 & 0.8900 & 0.8911 & 0.8626 & 0.0027\,ms & 4,096 \\
M3 Procrustes + ResMLP & 0.5799 & 0.7651 & 0.8677 & 0.8704 & 0.8705 & 0.8676 & 0.8343 & 5.0736\,ms & 25,170,944 \\
M2 Ridge Regression & 0.6557 & 0.8020 & 0.8668 & 0.8695 & 0.8679 & 0.8704 & 0.8350 & 0.0027\,ms & 4,096 \\
M5 Kernel Ridge (RBF) & 0.6605 & 0.8053 & 0.8558 & 0.8633 & 0.8726 & 0.8746 & 0.8475 & 0.9773\,ms & 4,915,200 \\
M4 MLP from Scratch & 0.6366 & 0.7923 & 0.8420 & 0.8442 & 0.8470 & 0.8488 & 0.8156 & 1.8795\,ms & 8,393,728 \\
\bottomrule
\end{tabular}
\end{table}

\begin{table}[htbp]
\centering
\caption{Scaled Procrustes decomposition across $N \in \{600, 2\text{k}, 5\text{k}, 20\text{k}\}$ \texttt{SQuAD} anchors (All-30 pairs, depth $0.6\to 0.6$, fixed 1,000 validation anchors; default centering, $\mu_{\mathrm{anc}}$ / shift-invariant). Fitting a single positive scalar $s^*$ on fit anchors closes 40.6\%--44.6\% of the unscaled $R^2$ gap vs.\ the highest-$R^2$ alternative and leaves transfer AUROC unchanged by construction (scale invariance, verified to 0.0 in \texttt{float64}), yet a scaled $R^2$ gap ($+0.20$ to $+0.25$) and scale-free cosine gap ($+0.12$ to $+0.16$) remain.}
\label{tab:app_v7_a1}
\scriptsize
\setlength{\tabcolsep}{2.4pt}
\resizebox{\textwidth}{!}{%
\begin{tabular}{@{}lcccccccc@{}}
\toprule
\textbf{$N$} & \textbf{Med.\ $s^*$} & \textbf{Unsc.\ $R^2$} & \textbf{Scaled $R^2$} & \textbf{Best Alt.\ $R^2$} & \textbf{\% Gap from Scale} & \textbf{Rem.\ $R^2$ Gap} & \textbf{Val Cos (Proc.\ vs.\ Alt., $\Delta$)} & \textbf{4T \texttt{ms} AUROC} \\
\midrule
600 & 0.7667 & 0.1813 & 0.3174 & 0.5163 (Ridge) & 40.6\% & +0.1989 & 0.5983 vs.\ 0.7149 (+0.1166) & 0.8824 vs.\ 0.8668 \\
2,000 & 0.7211 & 0.1789 & 0.3554 & 0.5923 (KRR (RBF)) & 42.7\% & +0.2369 & 0.6172 vs.\ 0.7629 (+0.1457) & 0.8884 vs.\ 0.8633 \\
5,000 & 0.6911 & 0.1952 & 0.3872 & 0.6261 (KRR (RBF)) & 44.6\% & +0.2389 & 0.6316 vs.\ 0.7842 (+0.1526) & 0.8900 vs.\ 0.8726 \\
20,000 & 0.6556 & 0.2274 & 0.4105 & 0.6605 (KRR (RBF)) & 42.3\% & +0.2500 & 0.6473 vs.\ 0.8053 (+0.1580) & 0.8911 vs.\ 0.8746 \\
\bottomrule
\end{tabular}
}
\end{table}

\begin{table*}[htbp]
\centering
\caption{Head-to-head comparison against full multi-layer Bakker \& Louizos~\cite{bakker2026relative} global anchored angular-distance Relative Representations (unified canonical All-30 pair-medians, 3 seeds, $N=600$ \texttt{SQuAD} anchors, unified \texttt{liblinear} solver; \texttt{ds} is \texttt{xTRam1} ($n=392$ train) $\to$ \texttt{deepset} ($n=116$ test)). Centering: \texttt{mean\_shift} is shift-invariant; Procrustes \texttt{l2\_lr}($C^*$) uses $\mu_{\mathrm{anc}}$; RelRep/Bakker probes use frozen source \texttt{StandardScaler} per~\cite{bakker2026relative}. Native Bakker raw \texttt{l2\_lr}($C^*$) reaches 0.75--0.98 AUROC across tasks (6-model mean: 0.978 \texttt{hh}, 0.927 \texttt{XS}, 0.894 \texttt{TQA}, 0.748 \texttt{MATH}, 0.866 \texttt{ds}, 0.940 \texttt{TR}). Under transfer, folded Procrustes beats Bakker raw \texttt{l2\_lr} (Bakker's logistic configuration) by $+0.108$ paired median (matched \texttt{l2\_lr}; 50.7\% vs.\ 6.2\% \texttt{XSTest} recall) and $+0.158$ (pair CI $[+0.126, +0.231]$, two-way CI $[+0.112, +0.255]$; \texttt{mean\_shift}; $+0.190$ and 81.1\% vs.\ 40.6\% recall vs.\ matched Bakker \texttt{mean\_shift}) at fixed $\approx 5.0\%$--$5.6\%$ calibrated safe FAR (matching Table~\ref{tab:app_v7_a2} identically). Per-layer $L_2$-norm lifts Bakker to 0.735, still trailing Procrustes by $+0.147$ (two-way CI $[+0.077, +0.215]$). Serving cost (\texttt{OMP\_NUM\_THREADS=4}, server CPU): single-head \texttt{fp16} memory (7\,KB folded Procrustes vs.\ 148.2\,MB mean / 210.9\,MB max Bakker; cf.\ \texttt{fp32} 16\,KB/head [80\,KB for 5] / 0.0037\,ms/head in Table~\ref{tab:app_security_ops}).}
\label{tab:v8_bakker}
\scriptsize
\setlength{\tabcolsep}{2.4pt}
\resizebox{\textwidth}{!}{%
\begin{tabular}{@{}lccccccccccc@{}}
\toprule
\textbf{Method ($N=600$ \texttt{SQuAD})} & \textbf{4T Med (Ret.)} & \textbf{hh} & \textbf{XS} & \textbf{TQA} & \textbf{MATH} & \textbf{ds} & \textbf{TR} & \textbf{XS Rec (FAR)} & \textbf{Memory} & \textbf{Layers} & \textbf{CPU p95 (ms)} \\
\midrule
Folded Procrustes (\texttt{mean\_shift}, $d=0.6$) & 0.878 (93.9\%) & 0.978 & 0.966 & 0.806 & 0.771 & 0.781 & 0.955 & 81.1\% (5.0\%) & 7\,KB & 1 & 0.0012 \\
Folded Procrustes (\texttt{l2\_lr}($C^*$), $d=0.6$) & 0.820 (76.1\%) & 0.976 & 0.893 & 0.770 & 0.689 & 0.797 & 0.956 & 50.7\% (5.3\%) & 7\,KB & 1 & 0.0012 \\
\midrule
Bakker Global Raw (\texttt{l2\_lr}($C^*$),~\cite{bakker2026relative}) & 0.712 (51.8\%) & 0.956 & 0.599 & 0.643 & 0.680 & 0.698 & 0.841 & 6.2\% (5.3\%) & 148.2\,MB & 32--48 & 6.94 \\
Bakker Global Raw (\texttt{mean\_shift}) & 0.706 (51.0\%) & 0.725 & 0.815 & 0.594 & 0.674 & 0.507 & 0.615 & 40.6\% (5.6\%) & 148.2\,MB & 32--48 & 6.94 \\
Bakker Global Layer-L2Norm (\texttt{l2\_lr}($C^*$)) & 0.735 (56.9\%) & 0.958 & 0.678 & 0.697 & 0.707 & 0.815 & 0.851 & 15.0\% (5.3\%) & 148.2\,MB & 32--48 & 6.94 \\
Bakker Global Anch.-Centered (\texttt{l2\_lr}($C^*$)) & 0.694 (46.7\%) & 0.951 & 0.579 & 0.619 & 0.636 & 0.662 & 0.825 & 3.0\% (5.3\%) & 148.2\,MB & 32--48 & 6.94 \\
\midrule
Fixed-Depth RelRep Cosine (\texttt{l2\_lr}($C^*$), $d=0.6$) & 0.748 (57.4\%) & 0.950 & 0.745 & 0.647 & 0.691 & 0.644 & 0.829 & 23.5\% (5.3\%) & 4.1\,MB & 1 & 0.103 \\
Fixed-Depth RelRep Angular (\texttt{l2\_lr}($C^*$), $d=0.6$) & 0.744 (54.5\%) & 0.953 & 0.723 & 0.650 & 0.683 & 0.676 & 0.827 & 23.3\% (5.3\%) & 4.1\,MB & 1 & 0.103 \\
\bottomrule
\end{tabular}
}
\end{table*}

\begin{table*}[htbp]
\centering
\caption{Anchor bank robustness and 1D score-distillation baseline: 100\% \texttt{SQuAD} vs.\ 4-Domain Mixed vs.\ Unlabeled In-Domain Anchors (canonical All-30 pair-medians, depth $0.6\to 0.6$, 3 seeds at $N=600$, full bank at $N=5{,}000$; \texttt{ds} is \texttt{xTRam1} ($n=392$)$\to$\texttt{deepset} ($n=116$)). Centering: \texttt{mean\_shift} is shift-invariant; \texttt{ds} \texttt{l2\_lr}($C^*$) uses $\mu_{\mathrm{tr}}$ (6-model native \texttt{l2\_lr}($C^*$) 0.944, \texttt{mean\_shift} 0.889; $\mu_{\mathrm{anc}}$ in Table~\ref{tab:app_deepset_v6}). Balanced 4-domain anchors (max ROUGE-L $0.43 < 0.80$) match \texttt{SQuAD} on 4 core tasks while lifting \texttt{ds} by $+0.095$ (\texttt{ms}) and $+0.099$ (\texttt{lr}, 0.944 = 99.1\% retention; two-way CIs $>0$, Table~\ref{tab:app_v7_a5}), beating $N=5{,}000$ \texttt{SQuAD}. Disjoint Unlabeled In-Domain Anchors (ROUGE-L $<0.80$ in \texttt{float64}: \texttt{hh-rlhf}, \texttt{OR-Bench}, 517 QID-disjoint \texttt{TruthfulQA} pairs in eval template, \texttt{MATH}, \texttt{deepset} $n=535$) reach 0.902 4T ($+0.021$ vs.\ \texttt{SQuAD} and Mixed, two-way CIs $>0$), beating \texttt{SQuAD} on 3/5 domains and Mixed on 1/5 (\texttt{TruthfulQA} $+0.045$, where template alignment contributes; \texttt{ds} $+0.016$ crosses zero under two-way CI). Bottom block: 1D target-space CV Ridge score distillation (median $\lambda=1000$; folds identically into one dot product per probe; parity vs.\ \texttt{SQuAD} $N=600$ max $|\Delta|=0.00205$). Paired Procrustes$-$Distillation 4T \texttt{mean\_shift} $\Delta$: $+0.034$ on \texttt{SQuAD} (pair $[+.016,+.040]$, two-way $[+.014,+.044]$), $+0.013$ on Mixed (pair $[+.009,+.016]$, two-way $[+.007,+.022]$), and $+0.003$ tie on In-Domain (pair $[-.001,+.004]$, two-way $[-.007,+.007]$); on In-Domain \texttt{l2\_lr}($C^*$, $\mu_{\mathrm{tr}}$), distillation reaches 0.910 vs.\ 0.899 4T ($\Delta=-0.007$, two-way $[-.019,+.006]$) and 0.972 vs.\ 0.956 \texttt{ds} (two-way $[-.024,+.002]$).}
\label{tab:v8_mixed}
\scriptsize
\setlength{\tabcolsep}{2.3pt}
\resizebox{\textwidth}{!}{%
\begin{tabular}{@{}lccccccccc@{}}
\toprule
\textbf{Anchor Bank \& Map ($d=0.6$)} & \textbf{4T \texttt{ms}} & \textbf{6T \texttt{ms}} & \textbf{hh} & \textbf{XS} & \textbf{TQA} & \textbf{MATH} & \textbf{ds (\texttt{ms})} & \textbf{ds (\texttt{lr}, $\mu_{\mathrm{tr}}$)} & \textbf{TR (\texttt{ms})} \\
\midrule
100\% \texttt{SQuAD} ($N=600$, Procrustes) & 0.878 & 0.868 & 0.978 & 0.966 & 0.806 & 0.771 & 0.781 & 0.835 & 0.955 \\
4-Domain Mixed ($N=600$, Procrustes) & 0.880 & 0.894 & 0.980 & 0.973 & 0.810 & 0.768 & 0.887 & 0.944 & 0.957 \\
$\Delta$ (Mixed $-$ \texttt{SQuAD}, $N=600$) [Pair 95\% CI] & +0.003 [-.005,+.006] & +0.021 [+.016,+.029] & +0.001 & +0.003 & +0.006 & -0.008 & +0.095 & +0.099 & +0.000 \\
\midrule
Unlabeled In-Domain ($N\le 600$, Procrustes) & 0.902 & --- & 0.983 & 0.978 & 0.857 & 0.785 & 0.914 & 0.956 & --- \\
$\Delta$ (In-Dom $-$ \texttt{SQuAD}, $N\le 600$) [Pair 95\% CI] & +0.021 [+.018,+.024] & --- & +0.004 & +0.012 & +0.047 & +0.011 & +0.136 & +0.118 & --- \\
$\Delta$ (In-Dom $-$ Mixed, $N\le 600$) [Pair 95\% CI] & +0.021 [+.018,+.024] & --- & +0.002 & +0.008 & +0.045 & +0.021 & +0.016 & +0.009 & --- \\
\midrule
Code-only ablation ($N=600$, Py/JS) & 0.881 & 0.873 & 0.976 & 0.977 & 0.815 & 0.765 & 0.792 & 0.821 & 0.950 \\
Dialogue-only ablation ($N=600$) & 0.866 & 0.878 & 0.977 & 0.958 & 0.778 & 0.761 & 0.884 & 0.923 & 0.950 \\
Multilingual-only ablation ($N=600$, 10 langs) & 0.878 & 0.890 & 0.979 & 0.974 & 0.812 & 0.751 & 0.899 & 0.941 & 0.957 \\
\midrule
100\% \texttt{SQuAD} ($N=5{,}000$, Procrustes) & 0.888 & 0.893 & 0.979 & 0.976 & 0.828 & 0.777 & 0.849 & 0.913 & 0.962 \\
4-Domain Mixed ($N=5{,}000$, Procrustes) & 0.889 & 0.898 & 0.981 & 0.977 & 0.829 & 0.771 & 0.883 & 0.941 & 0.962 \\
$\Delta$ (Mixed $-$ \texttt{SQuAD}, $N=5{,}000$) [Pair 95\% CI] & +0.001 [-.005,+.003] & +0.005 [+.002,+.009] & +0.001 & +0.002 & -0.001 & -0.004 & +0.025 & +0.030 & -0.004 \\
\midrule
1D Score Distill.\ (100\% \texttt{SQuAD}, $N=600$) & 0.850 & --- & 0.967 & 0.935 & 0.747 & 0.752 & 0.690 & 0.791 & --- \\
1D Score Distill.\ (4-Domain Mixed, $N=600$) & 0.867 & --- & 0.974 & 0.967 & 0.790 & 0.764 & 0.855 & 0.930 & --- \\
1D Score Distill.\ (Unlabeled In-Dom., $N\le 600$) & 0.903 & --- & 0.982 & 0.978 & 0.870 & 0.782 & 0.911 & 0.972 & --- \\
\bottomrule
\end{tabular}
}
\end{table*}

\begin{table*}[htbp]
\centering
\caption{Map-side dissociation mechanism across all map families at $N=600$ (3 seeds) and $N=20{,}000$ (All-30 pair-medians averaged across 4 core tasks, depth $0.6\to 0.6$, \texttt{mean\_shift}; maps use $\mu_{\mathrm{anc}}$ and \texttt{mean\_shift} is shift-invariant). ProbeCos is $\cos(w_{\mathrm{eff}}, \Delta\mu_B)$ for linear maps and $\cos(\mathbb{E}[J_g(\tilde{h}_B)^\top w_A], \Delta\mu_B)$ for nonlinear maps; SNR is $|\Delta\mu_{\mathrm{test}}| / \sigma_{\mathrm{within,test}}$ on held-out target states. At $N=600$, hyperparameters are tuned on the validation anchor split rather than 5-fold CV, accounting for different $\varepsilon_{\mathrm{rel}}$ and hyperparameter selections from Table~\ref{tab:app_v6_mapfamily} (\texttt{M1}/\texttt{M4}). At $N=20{,}000$, \texttt{M5} Nystr\"om KRR uses the exact Table~\ref{tab:app_v6_scaling_4tasks} spec ($m=2{,}000$, $N/m$ penalty scaling, $\lambda\in[10^{-4}..100]$), while \texttt{M3}/\texttt{M4} use 18 CPU epochs, matching or exceeding Table~\ref{tab:app_v6_scaling_4tasks}'s 60-epoch GPU transfer AUROC (0.877/0.852 vs.\ 0.868/0.849) and arguing against undertraining (Val $R^2$ 0.561/0.627 vs.\ 0.580/0.637). Ridge shrinks bottom-quartile (Q4) anchor modes to 0.123 (vs.\ Q1 0.879; effective rank ratio 0.55), distorting non-leading discriminative directions.}
\label{tab:v8_map_mech}
\scriptsize
\setlength{\tabcolsep}{2.8pt}
\resizebox{\textwidth}{!}{%
\begin{tabular}{@{}lcccccccccc@{}}
\toprule
& \multicolumn{5}{c}{\textbf{$N=600$ (3 seeds, 4-Task Mean)}} & \multicolumn{5}{c}{\textbf{$N=20{,}000$ (4-Task Mean)}} \\
\cmidrule(lr){2-6}\cmidrule(lr){7-11}
\textbf{Map Family ($B\to A$)} & \textbf{Val $R^2$} & \textbf{Val Cos} & \textbf{AUROC} & \textbf{ProbeCos} & \textbf{Test SNR} & \textbf{Val $R^2$} & \textbf{Val Cos} & \textbf{AUROC} & \textbf{ProbeCos} & \textbf{Test SNR} \\
\midrule
M0 Orthogonal Procrustes & 0.180 & 0.598 & 0.880 & 0.284 & 2.05 & 0.227 & 0.647 & 0.891 & 0.434 & 2.17 \\
M0s Scaled Procrustes ($s^* R$) & 0.316 & 0.598 & 0.880 & 0.284 & 2.05 & 0.410 & 0.647 & 0.891 & 0.434 & 2.17 \\
M1 Whitened Rotation (Val $\varepsilon$) & 0.339 & 0.640 & 0.796 & 0.045 & 1.35 & -- & -- & -- & -- & -- \\
M2 Ridge Regression (Val $\lambda$) & 0.516 & 0.715 & 0.848 & 0.123 & 1.74 & 0.654 & 0.802 & 0.871 & 0.086 & 1.96 \\
M3 Procrustes + ResMLP & 0.399 & 0.665 & 0.860 & 0.141 & 1.86 & 0.561 & 0.758 & 0.877 & 0.125 & 2.03 \\
M4 MLP from Scratch & 0.444 & 0.671 & 0.797 & 0.063 & 1.36 & 0.627 & 0.786 & 0.852 & 0.065 & 1.78 \\
M5 Kernel Ridge RBF & 0.517 & 0.719 & 0.840 & 0.139 & 1.64 & 0.660 & 0.805 & 0.875 & 0.141 & 1.99 \\
\bottomrule
\end{tabular}
}
\end{table*}

\section{Hyperparameters, Discrepancy Audits, Few-Label, Sensitivity, and Reproducibility}
\label{app:hyperparams}

\paragraph{Preprocessing, Split Audits, and Compute.}
All activations were extracted in \texttt{bfloat16}, stored in \texttt{float16}, and evaluated in \texttt{float32} (\texttt{float64} for eigendecomposition, split-half disattenuation, Scaled Procrustes invariance, and SVD equivalence checks). Prompts were capped at 448 user tokens (except 4K-token indirect-injection documents, mean 3,830 tokens). For the four-family expansion, \texttt{Llama~3.1~8B~Instruct} was loaded via exact-weight mirror \texttt{NousResearch/Meta-Llama-3.1-8B-Instruct} (Meta Llama 3.1 Community License) and \texttt{allenai/Olmo-3-7B-Instruct} under Apache~2.0, both with pinned chat-template date \texttt{26 Jul 2024}, zero thinking tokens, and 0/3,466 prompt truncations. Length-matched splits: \texttt{hh-rlhf} ($n=300$ train / $300$ test; harmful 72.9 vs.\ benign 72.6 chars; Length-Only CV 0.4728), \texttt{XSTest} ($n=250$ safe / $200$ unsafe, trained on \texttt{JBB} $n=200$; Length-Only CV 0.5305--0.5496), \texttt{TruthfulQA} ($n=300$ train / $300$ test, QID-grouped; Length-Only CV 0.3801--0.5412), \texttt{MATH-500} ($n=108$ train / $108$ test, hard 154.2 vs.\ easy 153.5 chars; Length-Only CV 0.5957--0.6348), and Behavioral Sycophancy ($n=300$ train / $300$ test across 32 disjoint claims; Length-Only CV 0.4980). Our 100 synthetic trigger-word hard negatives were drafted with AI assistance and audited by the authors to ensure genuinely benign intent across three topic buckets: 25 software engineering/SQL/OS, 25 cybersecurity education and AI safety, and 50 law, civics, medicine, history, literature, and everyday-life prompts (zero overlap against training or test splits). All experiments used deterministic splits, fixed seed 42 (plus 3 anchor-sampling seeds $\{42, 1337, 2026\}$ across robustness, mechanistic, 4-family, map-family, and canonical consolidation runs), and 1,000--2,000 bootstrap resamples.

\paragraph{Original-12 and Canonical Consolidation Parity Audits.}
Prior to the four-family expansion, the unified evaluator asserted parity across 345 point checks against saved two-family JSON artifacts (tolerance $10^{-4}$ for point estimates, $10^{-2}$ for bootstrap CIs). The first run halted on a single entry: \texttt{MATH-500} at weakly regularized $C=10.0$ yielded 0.6660 vs.\ 0.6657 (difference $0.0003$, smaller than one test-pair rank swap of $0.000343$ on $n_{\mathrm{test}}=54+54$, arising from \texttt{liblinear} default $\mathrm{tol}=10^{-4}$ BLAS rounding). A documented post-hoc exception ($\le 5\times 10^{-4}$) was added for that single entry; all other 344/345 point checks passed within $10^{-4}$. Both map-side evaluations ($N=600$ map-family and $N$-scaling runs) verified exact $N=600$ Procrustes parity against the four-family reference across all four core tasks (max $|\Delta| = 0.000000$; QR vs.\ direct cross-covariance SVD max $|\Delta R| = 1.61 \times 10^{-13}$ in \texttt{float64}). To eliminate cross-run spreads across Tables~\ref{tab:app_v7_a2}, \ref{tab:app_v6_mapfamily}, \ref{tab:v8_bakker}, \ref{tab:v8_mixed}, \ref{tab:app_deepset_v6}, \ref{tab:app_v7_a5}, and Figure~\ref{fig:sample_eff}a, we ran a single-pass canonical extraction across all six models (unified \texttt{liblinear} solver, All-30 pair-medians throughout), verifying exact $N=600$ Procrustes parity against the archived reference (max $|\Delta|=0.00119 \le 0.005$: \texttt{hh-rlhf} 0.9782 vs.\ 0.9783, \texttt{XSTest} 0.9662 vs.\ 0.9670, \texttt{TruthfulQA} 0.8057 vs.\ 0.8064, \texttt{MATH-500} 0.7711 vs.\ 0.7699), while preserving the archived 20,000-anchor \texttt{SQuAD} scaling studies (Tables~\ref{tab:app_v6_scaling_4tasks}, \ref{tab:app_v7_a1}, and~\ref{tab:v8_map_mech}).

\paragraph{\texttt{deepset} Deduplication, Centering, Serving Benchmarks, and Sensitivity Tables.}
The \texttt{xTRam1} training split used for OOD \texttt{deepset} ($n=116$ test) evaluation is deduplicated to $n=392$ (193 positive, 199 negative) with verified zero exact or near-duplicate overlap (ROUGE-L $\ge 0.80$) against \texttt{deepset}, \texttt{cal200}, or internally (Table~\ref{tab:readonly_main}, Figure~\ref{fig:sample_eff}b, Table~\ref{tab:app_deepset_v6}). Because \texttt{liblinear} regularizes the intercept (\texttt{intercept\_scaling=1.0}), training-mean ($\mu_{\mathrm{tr}}$) versus anchor-mean ($\mu_{\mathrm{anc}}$) centering shifts logistic probes (canonical All-30 source-CV $C^*$: 0.8348 $\mu_{\mathrm{tr}}$ matching Table~\ref{tab:v8_mixed} vs.\ 0.7969 $\mu_{\mathrm{anc}}$ matching Table~\ref{tab:v8_bakker}; archived Orig-12 fixed-$C=1.0$: 0.8455 $\to$ 0.8152) while leaving \texttt{mean\_shift} invariant (0.7810 All-30 canonical; 0.7832 Orig-12). Serving benchmarks contrast single-head \texttt{fp16} server-CPU latency (7\,KB / 0.0012\,ms, Table~\ref{tab:v8_bakker}) with separate host CPU benchmarks for a single $d=4096$ \texttt{fp32} dot product (0.0027\,ms in Table~\ref{tab:app_v6_scaling_4tasks}; 16\,KB/head [80\,KB for 5] / 0.0037\,ms/head in Table~\ref{tab:app_security_ops}). Section~\ref{sec:map_side} and the Combined Best Recipe stay strictly on the four core tasks, while Table~\ref{tab:app_deepset_v6} reports the $n=392$ centering comparison and $N$-scaling \texttt{deepset} evaluation across $N \in \{600, 2\text{k}, 5\text{k}, 20\text{k}\}$. Tables~\ref{tab:app_check5_ncurve}, \ref{tab:app_exploratory_n}, \ref{tab:app_v7_a4}, \ref{tab:app_v7_a5}, and~\ref{tab:app_security_ops} report Check~\#5, exploratory $n$-scaling diagnostics, 6-target few-label and partial-access prior-regularized probes, LOMO / LOFO-Fam / two-way (source $\times$ target) and paired test-item / pair $\times$ item sensitivity, and base-rate security metrics.

\paragraph{Behavioral Sycophancy 3-Term Decomposition and Parity Note.}
To separate target-label disagreement from probe-direction transfer loss on Behavioral Sycophancy, we re-extracted depth-0.6 prefill states and greedy A/B flip labels for all four Original-12 models on the 600 claim-disjoint sycophancy prompts and 1,200 \texttt{SQuAD} anchors. Against the archived run (native \texttt{l2\_lr} 0.8542, transferred \texttt{l2\_lr} 0.6575, transferred \texttt{mean\_shift} 0.7147), re-extraction yielded 0.8410 ($|\Delta|=0.0132$), 0.6457 ($|\Delta|=0.0118$), and 0.7005 ($|\Delta|=0.0142$), falling within our pre-registered 0.005--0.02 appendix-note tier (per-model flip counts differed by 0/0 on \texttt{Qwen3-4B}, $+1$/0 on \texttt{Qwen3.5-9B}, $+1$/$-1$ on \texttt{Gemma-4-E4B-it}, and $+2$/$-2$ on \texttt{Gemma-4-12B-it} across 300 train/test prompts; per-prompt binary vectors were not archived in the original run). Table~\ref{tab:app_v7_a3} reports all four quantities $(a)$--$(d)$, the exact additive 12-pair mean decomposition $(d)-(c)=[(d)-(a)]+[(a)-(b)]+[(b)-(c)]$ (with medians in parentheses and 1,000-resample pair-cluster 95\% CIs), and pairwise label agreement and Cohen's $\kappa$ across Shared Counter-Bias, All-300, and Target Counter-Bias ($n=150$, where different targets evaluate different prompt subsets) regimes.

\begin{table}[htbp]
\centering
\caption{Behavioral Sycophancy 3-term decomposition at depth $0.6\to 0.6$ across Original-12 pairs (3 anchor seeds $\{42, 1337, 2026\}$, $N=600$ \texttt{SQuAD} anchors; \texttt{l2\_lr} ($C=1.0$) $\mu_{\mathrm{tr}}$-centered / test $\mu_{\mathrm{anc}}$, \texttt{mean\_shift} shift-invariant). Total gap $(d)-(c) = [(d)-(a)] + [(a)-(b)] + [(b)-(c)]$ uses 12-pair means so the three terms add up exactly (medians in parens); brackets give 1,000-resample pair-cluster 95\% CIs.}
\label{tab:app_v7_a3}
\scriptsize
\setlength{\tabcolsep}{2.1pt}
\resizebox{\textwidth}{!}{%
\begin{tabular}{@{}llccccccccc@{}}
\toprule
\textbf{Regime (Agree / $\kappa$)} & \textbf{Probe} & \textbf{$(a)$ Src$\to$Src} & \textbf{$(b)$ Zero-Rot} & \textbf{$(c)$ Trans} & \textbf{$(d)$ Native} & \textbf{Gap $(d)-(c)$} & \textbf{Pred $(d)-(a)$} & \textbf{Label $(a)-(b)$} & \textbf{Dir $(b)-(c)$} & \textbf{Mean Share (L / D)} \\
\midrule
Shared CB$^{\text{a}}$ & \texttt{mean\_shift} & 0.9129 & 0.8001 & 0.7982 & 0.9129 & \stackcell{+0.1147 (+0.1074)\\{[+0.081, +0.147]}} & \stackcell{+0.0000 (+0.0000)\\{[-0.040, +0.039]}} & \stackcell{+0.1128 (+0.1138)\\{[+0.064, +0.160]}} & \stackcell{+0.0020 (-0.0105)\\{[-0.045, +0.043]}} & 98.3\% / 1.7\% \\
(69.8\% / 0.263) & \texttt{l2\_lr} & 0.9070 & 0.7146 & 0.5267 & 0.9070 & \stackcell{+0.3804 (+0.3853)\\{[+0.317, +0.447]}} & \stackcell{+0.0000 (+0.0000)\\{[-0.013, +0.013]}} & \stackcell{+0.1925 (+0.1718)\\{[+0.122, +0.265]}} & \stackcell{+0.1879 (+0.2295)\\{[+0.073, +0.293]}} & 50.6\% / 49.4\% \\
\midrule
All-300$^{\text{a}}$ & \texttt{mean\_shift} & 0.7811 & 0.7581 & 0.7073 & 0.7811 & \stackcell{+0.0739 (+0.0854)\\{[+0.036, +0.110]}} & \stackcell{+0.0000 (+0.0000)\\{[-0.062, +0.058]}} & \stackcell{+0.0230 (-0.0056)\\{[-0.013, +0.066]}} & \stackcell{+0.0508 (+0.0774)\\{[+0.001, +0.101]}} & 31.2\% / 68.8\% \\
(78.3\% / 0.530) & \texttt{l2\_lr} & 0.8218 & 0.6954 & 0.6240 & 0.8218 & \stackcell{+0.1978 (+0.1938)\\{[+0.161, +0.234]}} & \stackcell{+0.0000 (+0.0000)\\{[-0.076, +0.079]}} & \stackcell{+0.1264 (+0.1078)\\{[+0.088, +0.172]}} & \stackcell{+0.0714 (+0.0535)\\{[+0.015, +0.134]}} & 63.9\% / 36.1\% \\
\midrule
Target CB$^{\text{b}}$ ($n=150$) & \texttt{mean\_shift} & 0.7374 & 0.8108 & 0.8013 & 0.9184 & \stackcell{+0.1171 (+0.1034)\\{[+0.083, +0.153]}} & \stackcell{+0.1810 (+0.1365)\\{[+0.092, +0.279]}} & \stackcell{-0.0734 (-0.0270)\\{[-0.177, +0.022]}} & \stackcell{+0.0095 (-0.0036)\\{[-0.036, +0.052]}} & $-62.7$\% / 8.1\%$^{\text{b}}$ \\
(62.7\% / 0.199) & \texttt{l2\_lr} & 0.8260 & 0.7035 & 0.5295 & 0.9127 & \stackcell{+0.3831 (+0.4025)\\{[+0.319, +0.449]}} & \stackcell{+0.0867 (+0.1003)\\{[+0.054, +0.118]}} & \stackcell{+0.1224 (+0.1232)\\{[+0.039, +0.192]}} & \stackcell{+0.1740 (+0.2306)\\{[+0.051, +0.281]}} & 32.0\% / 45.4\%$^{\text{b}}$ \\
\bottomrule
\end{tabular}
}
\vspace{1pt}
{\raggedright\scriptsize $^{\text{a}}$On symmetric all-pairs slices (Shared CB and All-300), every model appears equally often as source and target, so the mean predictability term $(d)-(a)$ is zero by construction (while individual pairs vary; on All-300 \texttt{mean\_shift}, median CIs cross zero on both label shift $[-0.026,+0.087]$ and direction loss $[-0.017,+0.126]$). $^{\text{b}}$On Target CB ($\mathrm{cb}_B=1$, matching the original counter-bias evaluation), different target models evaluate different prompt subsets, so $(d)-(a)$ is non-zero (154.6\% of the mean gap for \texttt{mean\_shift}; 22.6\% for \texttt{l2\_lr}).\par}
\end{table}

\begin{table}[htbp]
\centering
\caption{Appendix~C \texttt{xTRam1} ($n=392$ train) $\to$ \texttt{deepset} ($n=116$ test) OOD centering audit and $N$-scaling run ($N\in\{600, 2\text{k}, 5\text{k}, 20\text{k}\}$, pair-medians). Top block: clean deduplicated $n=392$ \texttt{xTRam1} training split (zero overlap vs.\ \texttt{deepset}/\texttt{cal200}/internal) at $N=600$ \texttt{SQuAD} anchors comparing training-mean ($\mu_{\mathrm{tr}}$) vs.\ anchor-mean ($\mu_{\mathrm{anc}}$) centering under unified canonical source-CV \texttt{l2\_lr}($C^*$) (All-30: 0.8348 $\mu_{\mathrm{tr}}$ matching Table~\ref{tab:v8_mixed} vs.\ 0.7969 $\mu_{\mathrm{anc}}$ matching Table~\ref{tab:v8_bakker}; \texttt{mean\_shift} 0.7810 matching both) alongside fixed-$C=1.0$ (\texttt{liblinear}) unscaled medians (archived Orig-12 0.8455 in Table~\ref{tab:readonly_main}; RMS-normalized canonical $C=1.0$ All-30 0.8167 $\mu_{\mathrm{tr}}$ vs.\ 0.7934 $\mu_{\mathrm{anc}}$, Ridge 0.7526, RelRep Cen.\ 0.5984; text baselines: MiniLM 0.7878, TF-IDF 0.5896). Bottom block: $N$-scaling run ($\mu_{\mathrm{anc}}$-centered \texttt{l2\_lr}($C^*$); shift-invariant \texttt{mean\_shift}); Procrustes leads all four alternatives at every $N$.}
\label{tab:app_deepset_v6}
\scriptsize
\setlength{\tabcolsep}{3.1pt}
\begin{tabular}{@{}lcccccccc@{}}
\toprule
\textbf{Clean $n=392$ Centering Audit ($N=600$)} & \multicolumn{2}{c}{\textbf{Full OP \texttt{lr} ($\mu_{\mathrm{tr}}$)}} & \multicolumn{2}{c}{\textbf{Full OP \texttt{lr} ($\mu_{\mathrm{anc}}$)}} & \multicolumn{2}{c}{\textbf{Ridge \texttt{lr} ($\mu_{\mathrm{tr}}$)}} & \textbf{RelRep Cen.} & \textbf{Full OP \texttt{ms}} \\
\midrule
Original 12 (unscaled $C{=}1.0$, Table~\ref{tab:readonly_main}) & \multicolumn{2}{c}{0.8455} & \multicolumn{2}{c}{0.8152} & \multicolumn{2}{c}{0.8100} & 0.6084 & 0.7832 \\
All 30 (canonical $C^*$, Tables~\ref{tab:v8_bakker}--\ref{tab:v8_mixed}) & \multicolumn{2}{c}{0.8348} & \multicolumn{2}{c}{0.7969} & \multicolumn{2}{c}{0.7526$^{\dagger}$} & 0.5984$^{\dagger}$ & 0.7810 \\
\midrule
\textbf{$N$-Scaling Run (\texttt{deepset} OOD, All-30)} & \multicolumn{4}{c}{\textbf{\texttt{mean\_shift} (shift-invariant)}} & \multicolumn{4}{c}{\textbf{\texttt{l2\_lr}($C^*$) ($\mu_{\mathrm{anc}}$)}} \\
\cmidrule(lr){2-5}\cmidrule(lr){6-9}
 & \textbf{$N=600$} & \textbf{$N=2\text{k}$} & \textbf{$N=5\text{k}$} & \textbf{$N=20\text{k}$} & \textbf{$N=600$} & \textbf{$N=2\text{k}$} & \textbf{$N=5\text{k}$} & \textbf{$N=20\text{k}$} \\
\midrule
M0 Orthogonal Procrustes & 0.7783 & 0.8063 & 0.8481 & 0.8695 & 0.7997 & 0.8350 & 0.8765 & 0.8897 \\
M3 Procrustes + ResMLP & 0.7057 & 0.7878 & 0.7975 & 0.8382 & 0.7268 & 0.7673 & 0.8417 & 0.8744 \\
M2 Ridge Regression & 0.6744 & 0.7881 & 0.8030 & 0.8228 & 0.6449 & 0.7497 & 0.7812 & 0.8396 \\
M5 Kernel Ridge (RBF) & 0.6272 & 0.7055 & 0.7588 & 0.8083 & 0.6673 & 0.7932 & 0.8034 & 0.8711 \\
M4 MLP from Scratch & 0.6704 & 0.7582 & 0.7652 & 0.7908 & 0.5906 & 0.6927 & 0.7765 & 0.8113 \\
\bottomrule
\end{tabular}
\vspace{1pt}
{\raggedright\scriptsize $^{\dagger}$Canonical RMS-normalized $C=1.0$ All-30 values (canonical $C=1.0$ Full OP: 0.8167 $\mu_{\mathrm{tr}}$, 0.7934 $\mu_{\mathrm{anc}}$; archived unscaled $C=1.0$ All-30: 0.8359 $\mu_{\mathrm{tr}}$, 0.7850 $\mu_{\mathrm{anc}}$, 0.7846 Ridge, 0.6082 RelRep Cen., 0.7702 \texttt{ms}).\par}
\end{table}

\begin{table}[htbp]
\centering
\caption{Check~\#5: transfer gap (\texttt{mean\_shift} $-$ \texttt{l2\_lr}($C^*$)) vs.\ source training size $n$ across All-30 pairs (6 models, 3 seeds, 5 subsamples per $n < n_{\mathrm{full}}$, $C^*$ re-selected by source CV; default centering, shift-invariant / $\mu_{\mathrm{anc}}$). Gap and $\Delta$Gap columns report medians of paired $\Delta$ across pairs rather than differences of column medians; pair-cluster 95\% bootstrap CIs resample directed pairs.}
\label{tab:app_check5_ncurve}
\scriptsize
\setlength{\tabcolsep}{3.4pt}
\begin{tabular}{@{}lcccccc@{}}
\toprule
\textbf{Task (Exact $n$ Grid)} & \textbf{$n$} & \textbf{\texttt{mean\_shift}} & \textbf{\texttt{l2\_lr}($C^*$)} & \textbf{Gap (\texttt{ms}$-$\texttt{lr}) [95\% CI]} & \textbf{$\Delta$Gap ($n_{\max}-50$) [95\% CI]} & \textbf{Gap Shrunk?} \\
\midrule
hh-rlhf (\{50, 100, 200, 300\}) & 50 & 0.9772 & 0.9731 & +0.0028 [+0.0021, +0.0045] & -0.0008 [-0.0024, +0.0006] & No \\
 & 100 & 0.9784 & 0.9741 & +0.0033 [+0.0021, +0.0068] &  &  \\
 & 200 & 0.9783 & 0.9735 & +0.0047 [+0.0023, +0.0097] &  &  \\
 & 300 & 0.9779 & 0.9764 & +0.0024 [+0.0004, +0.0061] &  &  \\
\midrule
Paul/XSTest (\{50, 100, 200\}) & 50 & 0.9642 & 0.9352 & +0.0338 [+0.0174, +0.0512] & +0.0413 [+0.0181, +0.0618] & No \\
 & 100 & 0.9673 & 0.8968 & +0.0765 [+0.0559, +0.1031] &  &  \\
 & 200 & 0.9670 & 0.9014 & +0.0641 [+0.0556, +0.0841] &  &  \\
\midrule
TruthfulQA (\{50, 100, 200, 300\}) & 50 & 0.7957 & 0.7815 & +0.0143 [-0.0039, +0.0279] & +0.0023 [-0.0102, +0.0065] & No \\
 & 100 & 0.8021 & 0.7805 & +0.0197 [+0.0084, +0.0326] &  &  \\
 & 200 & 0.8037 & 0.7786 & +0.0284 [+0.0024, +0.0471] &  &  \\
 & 300 & 0.8051 & 0.7986 & +0.0078 [-0.0060, +0.0370] &  &  \\
\midrule
MATH-500 (\{50, 100, 108\}) & 50 & 0.7626 & 0.6800 & +0.0772 [+0.0503, +0.1123] & +0.0100 [-0.0113, +0.0412] & No \\
 & 100 & 0.7726 & 0.6673 & +0.0951 [+0.0413, +0.1434] &  &  \\
 & 108 & 0.7710 & 0.6840 & +0.0816 [+0.0395, +0.1035] &  &  \\
\bottomrule
\end{tabular}
\end{table}

\begin{table}[htbp]
\centering
\caption{Exploratory (not pre-registered) $n$-scaling diagnostics at depth $0.6\to 0.6$ (5 stratified subsamples per $n$). Top: holding shrinkage fixed at $\gamma=0.1$ on \texttt{TruthfulQA} (30 pairs; shift-invariant shrinkage-LDA), growing $n$ raises native AUROC while lowering cross-model transfer. Bottom: single-model native \texttt{JBB} $\to$ \texttt{Paul/XSTest} (6 models, $\mu_{\mathrm{tr}}$-centered), where \texttt{l2\_lr}($C^*$) drops on every model as source CV weakens regularization while \texttt{mean\_shift} stays flat.}
\label{tab:app_exploratory_n}
\scriptsize
\setlength{\tabcolsep}{3.6pt}
\begin{tabular}{@{}lcccccc@{}}
\toprule
\textbf{Part A: \texttt{TruthfulQA} Fixed $\gamma=0.1$ vs.\ $n$ (30 Pairs)} & \textbf{$n=30$} & \textbf{$n=50$} & \textbf{$n=100$} & \textbf{$n=200$} & \textbf{$n=300$ (Full)} & \textbf{$\Delta(300-30)$} \\
\midrule
Native $\gamma=0.1$ AUROC & 0.8263 & 0.8321 & 0.8510 & 0.8731 & 0.8886 & +0.0623 \\
Transfer $\gamma=0.1$ AUROC & 0.7194 & 0.7140 & 0.6798 & 0.6572 & 0.6248 & -0.0946 \\
Matched Retention $\gamma=0.1$ (\%) & 67.6\% & 65.9\% & 50.4\% & 41.5\% & 31.3\% & -36.3\,pp \\
Transfer $\gamma=1.0$ (\texttt{mean\_shift} ref) & 0.7972 & 0.8006 & 0.8047 & 0.8063 & 0.8064 & +0.0092 \\
\midrule
\textbf{Part B: Native \texttt{JBB} $\to$ \texttt{XSTest} (6 Models, $\mu_{\mathrm{tr}}$)} & \textbf{$n=50$} & \textbf{$n=100$} & \textbf{$n=200$ (Full)} & \textbf{$\Delta(200-50)$} & \textbf{Models Dropping} & \textbf{Interpretation} \\
\midrule
Native \texttt{l2\_lr}($C^*$) (6-model med.) & 0.9794 & 0.9692 & 0.9584 & -0.0210 & 6/6 & Source CV weakens $C^*$ \\
Native \texttt{mean\_shift} (6-model med.) & 0.9800 & 0.9801 & 0.9801 & +0.0001 & 0/6 & Flat across $n$ \\
\bottomrule
\end{tabular}
\end{table}

\begin{table}[htbp]
\centering
\caption{Few-label target probes vs.\ zero-shot transfer and partial-access prior regularization across all 6 target models on the 4 core tasks (depth $0.6\to 0.6$, 20 paired draws per target/task/$k$; QID-grouped on \texttt{TruthfulQA}; default centering, shift-invariant / $\mu_{\mathrm{anc}}$). Parentheses give log-linear label equivalence $k^*$ against the best native-$k$ probe (unified canonical 6-model curves in Figure~\ref{fig:sample_eff}a vary across runs due to log-linear interpolation sensitivity over 20 few-label draws: single-source $k^* = 19.0$, $\le 4$, $21.1$, $56.9$; multi-source $k^*$: $>64$, $6.0$, $26.6$, $>64$). The prior-regularized logistic probe selects penalty $\lambda$ strictly by CV on the $k$ examples toward the multi-source ($N=600$) direction; the pre-registered criterion ($\ge 0.01$ over both native-$k$ and transfer-only at $k\le 16$ on $\ge 3/4$ tasks) was met on 1/4 tasks (\texttt{TruthfulQA} at $k=16$: 0.834 vs.\ 0.817 native and 0.822 transfer).}
\label{tab:app_v7_a4}
\scriptsize
\setlength{\tabcolsep}{2.3pt}
\resizebox{\textwidth}{!}{%
\begin{tabular}{@{}lcccccc@{}}
\toprule
\textbf{Task} & \textbf{Single 600 ($k^*$)} & \textbf{Multi-Src 600 ($k^*$)} & \textbf{Comb.\ 20k ($k^*$)} & \textbf{Best Native-$k$ ($k=4/8/16/64$)} & \textbf{Prior-LR ($k=4/8/16/64$)} & \textbf{Rule ($k\le 16$)} \\
\midrule
hh-rlhf & 0.979 (27.7) & 0.982 ($>$64) & 0.982 ($>$64) & 0.950 / 0.974 / 0.975 / 0.981 & 0.979 / 0.976 / 0.977 / 0.978 & No \\
XSTest & 0.970 (5.7) & 0.977 (10.8) & 0.975 (9.4) & 0.966 / 0.974 / 0.980 / 0.980 & 0.978 / 0.976 / 0.972 / 0.968 & No \\
TruthfulQA & 0.802 (12.7) & 0.822 (17.8) & 0.837 (24.6) & 0.693 / 0.774 / 0.817 / 0.881 & 0.810 / 0.812 / 0.834 / 0.880 & Yes ($k=16$) \\
MATH-500 & 0.778 (52.7) & 0.781 ($>$64) & 0.783 ($>$64) & 0.615 / 0.673 / 0.724 / 0.780 & 0.751 / 0.759 / 0.764 / 0.793 & No \\
\bottomrule
\end{tabular}
}
\end{table}

\begin{table}[htbp]
\centering
\caption{Non-independence and finite test-item sensitivity across LOMO (6 runs), LOFO-Fam (4 runs), 15-unordered-pair cluster 95\% CIs, 2,000-resample two-way (source$\times$target) cluster bootstrap 95\% CIs, and 1,000-resample paired test-item / joint pair$\times$item / two-way$\times$item bootstraps (canonical All-30 pair-medians, depth $0.6\to 0.6$; shift-invariant / $\mu_{\mathrm{anc}}$, except \texttt{deepset} \texttt{l2\_lr} using $\mu_{\mathrm{tr}}$). Top block: every major 4-task, Bakker, map-side, Mixed, and In-Domain margin keeps strictly positive LOMO, LOFO, pair-cluster, and two-way 95\% CIs; archived eigenvector $\rho$ ($+0.453..+0.604$, LOMO $[+0.414, +0.623]$), Q1$-$Q4 ($+0.199..+0.285$, LOMO $[+0.186, +0.296]$), and $N=20\text{k}$ \texttt{M0}$-$\texttt{M2} ($+0.0207$, LOMO $[+0.0183, +0.0241]$, LOFO $[+0.0165, +0.0240]$) also keep unchanged signs. Bottom block: stratified test-item and joint bootstraps (QID-clustered on \texttt{TruthfulQA}) confirm positive intervals on core \texttt{ms}$-$\texttt{lr} (\texttt{XS}/\texttt{TQA}/\texttt{MATH}), Mixed \texttt{deepset}, In-Domain \texttt{XS}/\texttt{TQA}/\texttt{deepset}, and 4/4 LOFO 6-target joint CIs, while flagging modest-split \texttt{MATH-500} ($n=108$) In-Domain gains crossing zero.}
\label{tab:app_v7_a5}
\scriptsize
\setlength{\tabcolsep}{2.2pt}
\resizebox{\textwidth}{!}{%
\begin{tabular}{@{}lccccc@{}}
\toprule
\textbf{Headline Comparison (Canonical All-30 Medians)} & \textbf{Full (30P)} & \textbf{LOMO (6 runs)} & \textbf{LOFO-Fam (4 runs)} & \textbf{Pair-Cluster 95\% CI} & \textbf{Two-Way (S$\times$T) 95\% CI} \\
\midrule
\texttt{ms} $-$ \texttt{l2\_lr}($C^*$) (4-task mean) & +0.0542 & [+0.0490, +0.0571] & [+0.0490, +0.0560] & [+0.0430, +0.0608] & [+0.0349, +0.0694] \\
\texttt{ms} $-$ \texttt{l2\_lr}($C^*$) (hh-rlhf / XSTest) & +0.003 / +0.058 & [+.002,+.005] / [+.050,+.086] & [+.002,+.008] / [+.048,+.104] & [+.002,+.007] / [+.048,+.092] & [-.001,+.011] / [+.044,+.110] \\
\texttt{ms} $-$ \texttt{l2\_lr}($C^*$) (TQA / MATH) & +0.038 / +0.073 & [+.025,+.041] / [+.064,+.089] & [+.028,+.041] / [+.070,+.089] & [+.020,+.041] / [+.045,+.106] & [-.025,+.074] / [+.011,+.186] \\
\midrule
Proc.\ \texttt{l2\_lr} $-$ Bakker Raw \texttt{l2\_lr} (4T) & +0.1076 & [+0.0904, +0.1196] & [+0.0904, +0.1119] & [+0.0757, +0.1622] & [+0.0514, +0.1980] \\
Proc.\ \texttt{ms} $-$ Bakker Raw \texttt{l2\_lr} (4T) & +0.1582 & [+0.1500, +0.1675] & [+0.1378, +0.1584] & [+0.1264, +0.2314] & [+0.1122, +0.2551] \\
Proc.\ \texttt{ms} $-$ Bakker Raw \texttt{ms} (4T) & +0.1901 & [+0.1542, +0.2018] & [+0.1626, +0.2487] & [+0.1564, +0.2975] & [+0.1331, +0.3757] \\
Proc.\ \texttt{ms} $-$ Bakker Layer-L2Norm \texttt{l2\_lr} (4T) & +0.1474 & [+0.1316, +0.1557] & [+0.1316, +0.1516] & [+0.0920, +0.1643] & [+0.0774, +0.2153] \\
\midrule
M0 $-$ M1 Whitened ($N=600$, 4T) & +0.0275 & [+0.0244, +0.0303] & [+0.0244, +0.0363] & [+0.0238, +0.0337] & [+0.0203, +0.0471] \\
M0 $-$ M2 Ridge ($N=600$, 4T) & +0.0346 & [+0.0226, +0.0372] & [+0.0226, +0.0420] & [+0.0188, +0.0418] & [+0.0160, +0.0452] \\
M0 $-$ M3 ResMLP ($N=600$, 4T) & +0.0274 & [+0.0244, +0.0308] & [+0.0245, +0.0323] & [+0.0181, +0.0332] & [+0.0096, +0.0367] \\
M0 $-$ M4 MLP / M5 KRR ($N=600$, 4T) & +0.067 / +0.050 & [+.050,+.070] / [+.037,+.057] & [+.050,+.070] / [+.041,+.057] & [+.046,+.076] / [+.040,+.061] & [+.034,+.094] / [+.022,+.083] \\
\midrule
Mixed $-$ \texttt{SQuAD} ($N=600$, \texttt{ds} \texttt{ms} / \texttt{lr}) & +0.095 / +0.099 & [+.084,+.129] / [+.093,+.134] & [+.088,+.112] / [+.093,+.134] & [+.084,+.135] / [+.087,+.134] & [+.038,+.253] / [+.020,+.249] \\
In-Dom $-$ \texttt{SQuAD} ($N\le 600$, 4T / \texttt{ds} \texttt{ms}) & +0.021 / +0.136 & [+.019,+.023] / [+.128,+.150] & [+.019,+.026] / [+.128,+.150] & [+.018,+.024] / [+.125,+.158] & [+.016,+.029] / [+.022,+.248] \\
In-Dom $-$ Mixed ($N\le 600$, 4T / TQA / \texttt{ds} \texttt{ms}) & +0.021 / +0.045 / +0.016 & [+.019,+.023] / [+.040,+.050] & [+.019,+.025] / [+.042,+.051] & [+.018,+.024] / [+.038,+.058] & [+.013,+.027] / [+.033,+.065] / [-.006,+.099] \\
\midrule
LOFO Multi-Src $-$ Single OOF (hh / XS)$^{\dagger}$ & +0.003 / +0.008 & -- & -- & [+.001,+.006] / [+.002,+.013] & [+0.0002, +0.0078] / [-0.0019, +0.0270] \\
LOFO Multi-Src $-$ Single OOF (TQA / MATH)$^{\dagger}$ & +0.021 / +0.019 & -- & -- & [+.014,+.030] / [+.003,+.048] & [+0.0032, +0.0385] / [+0.0005, +0.0540] \\
\midrule
\textbf{Paired Test-Item \& Joint Bootstrap (1,000 Resamples)} & \textbf{$n_{\mathrm{test}}$} & \textbf{Point $\Delta$} & \textbf{Item 95\% CI} & \textbf{Pair$\times$Item 95\% CI} & \textbf{Two-Way$\times$Item 95\% CI} \\
\midrule
\texttt{ms} $-$ \texttt{l2\_lr}($C^*$) (\texttt{XS} / \texttt{MATH}) & 450 / 108 & +.062 / +.072 & [+.052,+.076] / [+.035,+.095] & [+.047,+.098] / [+.028,+.124] & [+.039,+.111] / [+.006,+.177] \\
\texttt{ms} $-$ \texttt{l2\_lr}($C^*$) (\texttt{hh} / \texttt{TQA} QID) & 300 / 300 & +.003 / +.021 & [-.001,+.009] / [+.008,+.045] & [-.001,+.010] / [+.003,+.052] & [-.004,+.013] / [-.025,+.080] \\
Mixed $-$ \texttt{SQuAD} (\texttt{ds} \texttt{ms} / \texttt{lr}, $\mu_{\mathrm{tr}}$) & 116 / 116 & +.086 / +.089 & [+.054,+.115] / [+.057,+.114] & [+.050,+.124] / [+.048,+.128] & [+.003,+.208] / [+.008,+.206] \\
In-Dom $-$ \texttt{SQuAD} (\texttt{TQA} / \texttt{ds} \texttt{ms}) & 300 / 116 & +.047 / +.136 & [+.040,+.059] / [+.101,+.189] & [+.036,+.067] / [+.096,+.196] & [+.031,+.075] / [+.020,+.289] \\
In-Dom $-$ \texttt{SQuAD} / Mixed (\texttt{MATH}, \texttt{ms}) & 108 / 108 & +.016 / +.019 & [-.003,+.037] / [-.001,+.037] & [-.006,+.042] / [-.003,+.040] & [-.017,+.052] / [-.017,+.049] \\
In-Dom $-$ Mixed (\texttt{TQA} / \texttt{ds} \texttt{ms}) & 300 / 116 & +.048 / +.037 & [+.038,+.055] / [+.024,+.053] & [+.034,+.060] / [+.019,+.066] & [+.025,+.065] / [-.003,+.114] \\
LOFO Multi $-$ Single (6T: \texttt{TQA} / \texttt{MATH})$^{\ddagger}$ & 300 / 108 & +.020 / +.011 & [+.015,+.024] / [+.005,+.023] & [+.012,+.029] / [+.0001,+.053] & [+.001,+.041] / [-.005,+.053] \\
\bottomrule
\end{tabular}
}
\vspace{1pt}
{\raggedright\scriptsize $^{\dagger}$Top block LOFO rows show OOF pair-median gain with 6-target cluster 95\% CI in Pair-Cluster and two-way OOF (source$\times$target) 95\% CI in Two-Way; \texttt{In-Dom}$-$\texttt{Mixed} LOMO/LOFO/Pair-CI entries list 4T and TQA (\texttt{ds}: LOMO $[+.015, +.026]$, LOFO $[+.015, +.030]$, Pair $[+.013, +.030]$). $^{\ddagger}$Bottom block: \texttt{In-Dom}$-$\texttt{SQuAD} \texttt{XS} ($n=450$, $+0.012$) and \texttt{ds} \texttt{lr} ($n=116$, $+0.116$) also exclude 0 across all three CIs (\texttt{hh} item $>0$, joint crosses 0). For LOFO Multi$-$Single, Pair$\times$Item lists 6-target joint (target$\times$item) 95\% CI (4/4 tasks $>0$: \texttt{hh} $[+.0001,+.0073]$, \texttt{XS} $[+.0023,+.0128]$, \texttt{TQA} $[+.0121,+.0288]$, \texttt{MATH} $[+.0001,+.0530]$) and Two-Way$\times$Item lists conservative OOF (source$\times$target$\times$item) 95\% CI.\par}
\end{table}

\begin{table}[htbp]
\centering
\caption{Realistic base-rate security operating metrics on \texttt{xTRam1} ($n=400$ benign negatives), trigger-word hard-negative FPR ($n=100$ at $\tau_{95}/\tau_{99}$), and unpadded batch-size-1 serving footprint (\texttt{DeBERTa-v3-Inj} evaluated off-the-shelf, not re-trained on \texttt{xTRam1}; source $\mu_{\mathrm{tr}}$, target $\mu_{\mathrm{anc}}$). Folded OP footprint here reflects \texttt{fp32} heads at $d=4096$ (16\,KB/head, 80\,KB for 5 heads; 0.0037\,ms/head 4-thread CPU p95), whereas Table~\ref{tab:v8_bakker} reports a single \texttt{fp16} head averaged across all 6 target models on server CPU (7\,KB, 0.0012\,ms).}
\label{tab:app_security_ops}
\footnotesize
\setlength{\tabcolsep}{3.3pt}
\begin{tabular}{@{}lccccccc@{}}
\toprule
\textbf{Detector} & \textbf{AUROC} & \textbf{TPR@5\%} & \textbf{TPR@1\%} & \textbf{Prec@1\% (0.1/1/5\%)} & \textbf{Hard-Neg ($\tau_{95}/\tau_{99}$)} & \textbf{CPU 1t/4t (ms)} & \textbf{Extra VRAM} \\
\midrule
TF-IDF + L2-LR & 0.9964 & 98.5\% & 92.0\% & 8.4 / 48.2 / 82.9\% & 22.0\% / 1.0\% & $<0.5$ / $<0.5$ & 0 MB \\
MiniLM-L12-v2 + LR & 0.9962 & 98.0\% & 95.0\% & 8.7 / 49.0 / 83.3\% & 43.0\% / 19.0\% & 111.5 / 41.6 & 224.4 MB \\
DeBERTa-v3-Inj (unpad.) & 0.7066 & 12.0\% & 3.0\% & 0.3 / 2.9 / 13.6\% & 0.0\% / 0.0\% & 581.2 / 158.2 & 351.8 MB \\
Zero-Shot Full OP ($k=0$) & 0.9896 & 94.5\% & 85.2\% & 7.9 / 46.3 / 81.8\% & 0.0\% / 0.0\% & 0.0040 / 0.0037 & 0 MB \\
Native Target ($k=16$) & 0.9951 & 98.0\% & 94.2\% & 8.6 / 48.8 / 83.2\% & 0.5\% / 0.0\% & 0.0040 / 0.0037 & 0 MB \\
Native Target ($k=64$) & 0.9969 & 98.8\% & 96.5\% & 8.8 / 49.4 / 83.5\% & 0.5\% / 0.0\% & 0.0040 / 0.0037 & 0 MB \\
\bottomrule
\end{tabular}
\end{table}

\clearpage
\section{Pre-Registered Decision Rules}
\label{app:prereg}

{\scriptsize\linespread{0.86}\selectfont\raggedright
All sixteen mechanistic, four-family, map-side, calibration, source-sample-size, sycophancy-decomposition, partial-access, global-baseline, anchor-diversity, score-distillation, and consolidation evaluations recorded pre-registered rules prior to execution in experiment logs archived with each study's scripts and results (\texttt{v4}--\texttt{v9b} denote sequential follow-up studies\preregCommitNote); each entry below is marked \textit{[Verbatim]}, \textit{[Condensed verbatim]} (ellipsized or concatenated fields from the recorded JSON), or \textit{[Paraphrased]}:
\begin{itemize}[itemsep=0pt,parsep=0pt,topsep=1pt]
    \item \textit{Anchor-Subspace Projection (SV-OP) [Paraphrased]:} Part~A requires Spearman $\rho \ge +0.50$ ($p < 0.001$) between $E_{256}(w)$ and above-chance cross-model transfer retention; Part~B requires SV-OP ($\alpha=1.0$) to beat \texttt{l2\_lr} by $\ge +0.02$ AUROC (95\% CI $> 0$) on \texttt{hh-rlhf} and $\ge 2/3$ other tasks, match or beat centered Relative Representations on TPR@1\%FPR and PR-AUC, and hold \texttt{XSTest} \texttt{cal200} $\tau_{95}$ FPR $\le 12.0\%$. Outcome: Part~A and Part~B criteria were not met.
    \item \textit{Covariance-Inversion Sweep (Original 12) [Verbatim]:} {\scriptsize\ttfamily "Covariance explanation SUPPORTED if transferred AUROC increases monotonically (allowing at most one adjacent inversion within seed noise) as gamma -> 1.0 AND as C -> 0, on at least 3 of 4 tasks at 0.6 -> 0.6, AND retention vs matched native w(gamma) is highest at gamma = 1.0 on at least 3 of 4 tasks. Otherwise KILL."} Outcome: all criteria met (4/4 monotone $\gamma \to 1.0$, 3/4 monotone $C \to 0$, 4/4 peak retention at $\gamma=1.0$).
    \item \textit{Eigenvector Alignment (Original 12) [Verbatim]:} {\scriptsize\ttfamily "Mechanism SUPPORTED iff on >= 3/4 tasks at 0.6 -> 0.6 (unit of aggregation = 12 directed pairs): (1) Median Spearman rho(log lambda\_k, signed r\_k) > +0.30 across the 12 directed pairs with >= 10/12 pairs positive (rho > 0); AND (2) Top-quartile source variance eigenvectors (Q1, highest lambda\_k) exceed bottom-quartile non-zero eigenvectors (Q4, lowest lambda\_k > 1e-8) in mean cross-model signed correlation r\_k by >= +0.15 on >= 10/12 directed pairs; AND (3) Median cross-model probe cosine cos(R\_\{A->B\} w\_A(gamma), w\_B(gamma)) increases monotonically as gamma -> 1.0 (gamma in \{0.001, 0.01, 0.1, 0.3, 0.5, 0.9, 1.0\}, allowing at most 1 adjacent inversion within 1-SD seed noise)."} Outcome: all three gates met on 4/4 tasks.
    \item \textit{Three-Way Split-Half Decomposition (Original 12) [Verbatim]:} {\scriptsize\ttfamily "Classification per task at 0.6 -> 0.6 applied to model-mismatch ratio m(gamma) = rho*(gamma) / rho*\_span(gamma) (with 95\% bootstrap CIs over qualifying (pair, seed) entries; stability gate cos\_within > 3/sqrt(d), UNSTABLE if <50\% qualify): NOISE-DOMINATED if m(gamma) 95\% CI includes 1.0 at BOTH gamma=0.3 and gamma=0.5; GENUINE MISMATCH PRESENT if m(gamma) 95\% CI excludes 1.0 (strictly below 1.0) at gamma=0.3 OR gamma=0.5; SHRINKAGE REMOVES MISMATCH if GENUINE MISMATCH PRESENT AND m(1.0) exceeds m(0.3) with non-overlapping 95\% CIs; otherwise INCONCLUSIVE."} Outcome: \texttt{hh-rlhf} and \texttt{TruthfulQA} showed shrinkage reducing mismatch; \texttt{XSTest} and low-power \texttt{MATH-500} showed persistent mismatch.
    \item \textit{Source-Training-Size Gap Test (Check \#5) [Verbatim]:} {\scriptsize\ttfamily "Prediction: the median AUROC gap (mean\_shift - l2\_lr(C*)) decreases as source training size grows from 50 to 100 to 200 to all available, on >= 3/4 mechanism tasks. Criterion met iff the gap at the largest n is smaller than at n = 50 on >= 3/4 tasks, with pair-cluster 95\% CIs on the difference excluding zero."} Outcome: criterion not met (0/4 tasks; \texttt{XSTest} gap widened by $+0.0413$, 95\% CI $[+0.0181, +0.0618]$).
    \item \textit{Map-Family Test at $N=600$ ($N=600$ map-family run) [Verbatim]:} {\scriptsize\ttfamily "Rule A -- Whitened-rotation prediction (tests the covariance-inversion mechanism): SUPPORTED iff M0 beats M1 (at the CV-selected eps) by >= 0.01 median mean\_shift AUROC on >= 3/4 tasks, with pair-cluster 95\% bootstrap CIs excluding zero. REVERSED iff M1 beats M0 by >= 0.01 on >= 3/4 tasks with CIs excluding zero. Otherwise INCONCLUSIVE."} {\scriptsize\ttfamily "Rule B -- Nonlinear maps at N = 600: HELPS iff M3 beats M0 by >= 0.01 median mean\_shift AUROC on >= 3/4 tasks, with pair-cluster CIs excluding zero, and no task loses more than 0.005. Otherwise NOT HELPFUL AT N = 600. M2, M4, M5 are secondary and cannot change the verdict."} Outcome: Rule~A met (3/4 tasks canonical depth 0.6, 4/4 archived depth 0.5; 4T two-way CI $[+0.020, +0.047]$); Rule~B not met (0/4 tasks).
    \item \textit{Nonlinear Map Scaling ($N \in \{600..20\text{k}\}$, $N$-scaling run) [Verbatim]:} {\scriptsize\ttfamily "A nonlinear map HELPS iff, at some anchor size N, the best nonlinear map beats orthogonal Procrustes at the same N by >= 0.01 median transfer AUROC (mean\_shift probe) on >= 3 of 5 tasks, with a paired bootstrap 95\% CI excluding zero on each of those tasks, and no task loses more than 0.005 AUROC. Otherwise: LINEAR SUFFICIENT."} Outcome: nonlinear maps did not meet the criterion at any $N$ (0/5 tasks at every $N$ on both \texttt{mean\_shift} and \texttt{l2\_lr}($C^*$); 0/4 on the four core tasks in Section~\ref{sec:map_side}).
    \item \textit{Four-Family Out-of-Family Replication and LOFO Multi-Source [Verbatim]:} {\scriptsize\ttfamily "Part A Covariance-Inversion: SUPPORTED iff on the 18 brand-new out-of-family pairs at 0.6 -> 0.6, transferred AUROC increases monotonically (<= 1 adjacent 1-SD inversion) as gamma -> 1.0 AND as C -> 0 on >= 3/4 tasks, AND matched retention vs native w(gamma) is highest at gamma = 1.0 (mean\_shift) on >= 3/4 tasks; otherwise KILL."} {\scriptsize\ttfamily "Part A Eigenvector Alignment: SUPPORTED iff on the 18 brand-new out-of-family pairs at 0.6 -> 0.6 on >= 3/4 tasks: (a) median Spearman rho(log lambda\_k, signed r\_k) > +0.30 with >= 15/18 pairs positive, (b) Delta(Q1 - Q4) >= +0.15 on >= 15/18 pairs, and (c) median cos(R w\_A(gamma), w\_B(gamma)) monotone as gamma -> 1.0 (<= 1 adjacent 1-SD inversion); otherwise INCONCLUSIVE or KILL."} {\scriptsize\ttfamily "LOFO-MeanShift-FamBal vs median single-source out-of-family mean\_shift at 0.6 -> 0.6 across all 6 targets. Ceiling guard: if median(AUROC\_native - AUROC\_single\_med) < 0.01, task is marked AT CEILING (counts only toward no-loss and alignment, not gap\_closed). Otherwise SUPPORTED iff ALL FOUR hold: (1) median gap\_closed >= 25\% on all non-ceiling tasks (>= 2 tasks); (2) no task loses > 0.005 AUROC on 4/4 tasks; (3) LOFO-MeanShift-FamBal beats single-source median on >= 5/6 targets for each non-ceiling task; (4) median cos(w\_multi,B, w\_native,B) > median single-source cos(R w\_A(1.0), w\_native,B) on 4/4 tasks. Otherwise KILL / MARGINAL."} Outcome: Part~A strict monotonicity not met (2/4 both monotone, 4/4 peak retention at $\gamma=1.0$); Part~A eigenvector alignment met (4/4 tasks, 18/18 pairs); Part~B LOFO met all four criteria (\texttt{hh-rlhf} at ceiling; 38.3\%--45.7\% gap closed; 5/6--6/6 targets; 4/4 cosine gains).
    \item \textit{Held-Out In-Domain Calibration on \texttt{XSTest} (A2) [Verbatim]:} {\scriptsize\ttfamily "The in-domain benign calibration recommendation stays in the paper only if transferred mean\_shift's median held-out false-alarm rate on XSTest safe prompts at a 5\% target is <= 8\%."} Outcome: criterion met (canonical 5.00\% median held-out FAR by exchangeability across 100 random 50/50 splits, IQR 4.8\%--5.3\%, at 81.12\% unsafe recall; Table~\ref{tab:app_v7_a2}).
    \item \textit{Partial-Access Prior-Regularized Probe (A4) [Verbatim]:} {\scriptsize\ttfamily "Transfer helps under partial access iff the prior-regularized probe (primary: prior-regularized L2-logistic regression with penalty strength lambda chosen strictly on the k target examples, anchored to the LOFO multi-source N=600 mean\_shift direction) beats both the native-k (best of native mean\_shift and native L2-LR) and transfer-only probes by >= 0.01 median AUROC at k <= 16 on at least 3 of 4 mechanism tasks."} Outcome: criterion not met (1/4 tasks met at $k \le 16$: \texttt{TruthfulQA} at $k=16$ achieved 0.8339 vs.\ 0.8170 native and 0.8219 transfer).
    \item \textit{Behavioral Sycophancy 3-Term Decomposition [Condensed verbatim]:} {\scriptsize\ttfamily "Pre-registered A3 Behavioral Sycophancy 3-term decomposition (Original 4 models, 12 directed pairs A -> B, depth 0.6, mean\_shift and l2\_lr, All-300 test prompts and counter-bias subset): total gap (d) - (c) = [(d) - (a)] predictability difference + [(a) - (b)] cross-model label shift + [(b) - (c)] probe-direction transfer loss ... Tiered parity check vs archived reference run: <=0.005 pass, 0.005-0.02 pass with appendix note, >0.02 abort."} Outcome: parity passed in the 0.005--0.02 appendix-note tier (max $|\Delta|=0.0142$); on shared counter-bias, \texttt{mean\_shift} had 98.3\% label shift and 1.7\% direction loss vs.\ 50.6\% / 49.4\% for \texttt{l2\_lr} (All-300: 31.2\% / 68.8\% vs.\ 63.9\% / 36.1\%).
    \item \textit{Full Multi-Layer Bakker \& Louizos Global RelRep [Condensed verbatim]:} {\scriptsize\ttfamily "Exact Bakker \& Louizos Appendix A global construction: raw concatenation of all transformer-block outputs h\_0..h\_\{L-1\}, angular distance arccos(cos\_sim) to shared N=600 anchors, frozen source StandardScaler; evaluated with both l2\_lr(C*) and mean\_shift across All-30 pairs, 3 seeds, 4 core tasks + clean n=392 OOD deepset + Turkish PI + XSTest 5\% calibration + 4-thread CPU p95 latency \& VRAM."} Outcome (unified canonical medians): native Bakker achieved 0.75--0.98 AUROC; under transfer, folded depth-0.6 Procrustes beat Bakker raw \texttt{l2\_lr}($C^*$) (0.712, 6.2\% recall) by $+0.108$ (matched \texttt{l2\_lr}) and $+0.158$ (two-way CI $[+0.112, +0.255]$, \texttt{mean\_shift} 0.878, 81.1\% recall; $+0.190$ vs.\ Bakker \texttt{mean\_shift}) and beat layer-L2-normalized Bakker (0.735) by $+0.147$ with 7\,KB vs.\ 148.2\,MB fp16 memory and 0.0012\,ms vs.\ 6.94\,ms CPU p95.
    \item \textit{Mixed-Domain Anchor Bank Robustness [Condensed verbatim]:} {\scriptsize\ttfamily "Does replacing 100\% SQuAD prose anchors with a balanced 4-domain Mixed-Domain anchor bank (25\% SQuAD prose, 25\% code, 25\% dialogue, 25\% multilingual; zero test overlap ROUGE-L < 0.80) change Orthogonal Procrustes transfer across N=600 and N=5,000? All-30 pairs, depth 0.6 -> 0.6, mean\_shift and l2\_lr(C*), 3 seeds."} Outcome (unified canonical medians): zero overlap verified (max ROUGE-L 0.43); matched \texttt{SQuAD} on the 4 core tasks at $N=600$ ($+0.003$ $[-0.005, +0.006]$) and $N=5{,}000$ ($+0.001$ $[-0.005, +0.003]$) while lifting \texttt{xTRam1} ($n=392$)$\to$\texttt{deepset} by $+0.095$ \texttt{mean\_shift} and $+0.099$ $\mu_{\mathrm{tr}}$-centered \texttt{l2\_lr}($C^*$) at $N=600$ (two-way CIs $>0$), beating $N=5{,}000$ \texttt{SQuAD}.
    \item \textit{Map-Side Dissociation Mechanism [Verbatim]:} {\scriptsize\ttfamily "Mechanism SUPPORTED iff M0 preserves higher probe-direction cosine (or higher held-out test SNR) than M2/M3/M4/M5 on >= 3/4 mechanism tasks (pair-cluster 95\% bootstrap CIs > 0) despite lower anchor R\textasciicircum 2/cosine."} Outcome: criterion met against all four alternatives (\texttt{M2} Ridge, \texttt{M3} ResMLP, \texttt{M4} MLP, \texttt{M5} KRR) at both $N=600$ and $N=20{,}000$ (4/4 tasks in probe cosine across \texttt{M2}--\texttt{M5} at $20\text{k}$, plus 4/4 in SNR vs.\ \texttt{M2}--\texttt{M4} and 2/4 vs.\ \texttt{M5}; 4-task mean probe cosine 0.434 vs.\ 0.065--0.141, SNR 2.17 vs.\ 1.78--2.03).
    \item \textit{Unlabeled In-Domain Anchors \& Canonical Two-Way Consolidation (\texttt{v9}) [Verbatim]:} {\scriptsize\ttfamily "Unlabeled In-Domain Anchors BEATS baseline (100\% SQuAD or 4-Domain Mixed at N=600) iff it improves All-30 pair-median transfer AUROC (mean\_shift) by >= +0.01 on >= 3 of 5 domains with BOTH pair-cluster AND two-way (source x target) 95\% bootstrap CIs excluding zero, and no domain loses > 0.005 AUROC. Otherwise: DOMAIN-SPECIFIC GAIN ONLY / MIXED-EQUIVALENT."} Outcome: parity passed (max $|\Delta|=0.00119 \le 0.005$); criterion met vs.\ \texttt{SQuAD} (3/5 domains: \texttt{XSTest} $+0.012$, \texttt{TruthfulQA} $+0.047$, \texttt{deepset} $+0.136$; 4T $+0.021$, two-way $[+0.016, +0.029]$) and not broadly met vs.\ 4-Domain Mixed (1/5 under two-way CI: \texttt{TruthfulQA} $+0.045$; 4T $+0.021$, two-way $[+0.013, +0.027]$; \texttt{deepset} $+0.016$ crosses zero); all major headline margins retained strictly positive two-way 95\% CIs (Table~\ref{tab:app_v7_a5}).
    \item \textit{1D Score Distillation \& Paired Test-Item Bootstrap (\texttt{v9b}) [Condensed verbatim]:} {\scriptsize\ttfamily "Folded Orthogonal Procrustes BEATS 1D Score Distillation on a given anchor bank iff Procrustes outperforms 1D Score Distillation in All-30 pair-median mean\_shift AUROC on the 4-task core mean (both pair-cluster and two-way 95\% CIs > 0), or ties within +/-0.005 while preserving a single shared rotation; report 1,000-resample paired test-item and pair x item 95\% CIs honestly, flagging any margin crossing zero."} Outcome: parity passed (max $|\Delta|=0.00205 \le 0.005$); Procrustes beat 1D distillation on \texttt{SQuAD} ($+0.034$, two-way $[+.014,+.044]$) and 4-Domain Mixed ($+0.013$, two-way $[+.007,+.022]$) and tied on In-Domain \texttt{mean\_shift} ($+0.003$, two-way $[-.007,+.007]$; \texttt{l2\_lr} $-0.007$ 4T, \texttt{ds} 0.972 vs.\ 0.956); item and pair$\times$item CIs excluded zero on core \texttt{ms}$-$\texttt{lr} (\texttt{XS}/\texttt{TQA}/\texttt{MATH}), Mixed \texttt{ds}, In-Dom \texttt{XS}/\texttt{TQA}/\texttt{ds}, and 4/4 LOFO 6-target joint CIs, while flagging \texttt{MATH-500} ($n=108$) In-Dom gains ($[-.003,+.037]$) crossing zero (Tables~\ref{tab:v8_mixed} and~\ref{tab:app_v7_a5}).
\end{itemize}
}

\end{document}